\documentclass[runningheads]{llncs}
\usepackage{amsmath, amssymb}
\usepackage{mathtools}
\usepackage{graphicx}
\usepackage{subcaption}
\usepackage[dvipsnames]{xcolor}
\usepackage[colorlinks=true,urlcolor=black, citecolor=black, linkcolor=black]{hyperref}
\usepackage{orcidlink}

\usepackage{cleveref}
\usepackage{tikz}
\usepackage{booktabs}

\usetikzlibrary{calc,matrix,positioning,patterns,shapes.geometric}
\newcommand{\citep}[1]{\cite{#1}}
\newcommand{\citet}[1]{\cite{#1}}
\usepackage{bm}
\usepackage{pifont}
\newcommand{\cmark}{\ding{51}}%
\newcommand{\xmark}{\ding{55}}%
\usepackage{flushend}
\usepackage{multirow}
\usepackage{xspace}

\newcommand{\bdelta}[0]{\bm{\delta}}

\def\ve{{\bm{e}}}
\def\vf{{\bm{f}}}

\def\vp{{\bm{p}}}

\def\vs{{\bm{s}}}

\def\vx{{\bm{x}}}
\def\vy{{\bm{y}}}

\def\vdelta{{\bm{\delta}}}
\def\vbeta{{\bm{\beta}}}

\def\mS{{\bm{S}}}

\def\mX{{\bm{X}}}

\DeclareMathOperator*{\argmax}{arg\,max}

\newcommand{\counter}{counterfactual\xspace}

\newcommand{\counterexps}{counterfactual explanations\xspace}

\newcommand{\advex}{adversarial example\xspace}
\newcommand{\advexs}{adversarial examples\xspace}

\crefname{condition}{condition}{conditions}
\Crefname{condition}{Condition}{Conditions}
\crefname{example}{example}{example}
\Crefname{example}{Example}{Example}
\Crefname{section}{Section}{Section} 
\crefname{section}{Sec.}{Sec.} 
\crefname{figure}{Fig.}{Figs.} 
\Crefname{figure}{Figure}{Figures} 
\Crefname{table}{Table}{Tables} 
\crefname{table}{Tab.}{Tab.} 
\Crefname{equation}{Equation}{Equations} 
\crefname{equation}{Eqn.}{Eqns.} 

\begin{document}
\title{Towards Non-Adversarial Algorithmic Recourse}
%
%
\author{Tobias Leemann\inst{1,2}\orcidlink{0000-0001-9333-228X}\and
Martin Pawelczyk\inst{3}\orcidlink{0000-0002-6191-4434} \and
Bardh Prenkaj\inst{2}\orcidlink{0000-0002-2991-2279} \and
Gjergji Kasneci\inst{2}\orcidlink{0000-0002-3123-7268}
}
\authorrunning{T. Leemann et al.}
%
\institute{University of Tübingen, Tübingen, Germany\\
\email{tobias.leemann@uni-tuebingen.de}
\and
Technical University of Munich, Munich, Germany
\and
Harvard University, Cambridge, MA, USA\\}

\maketitle              
\begin{abstract}
The streams of research on adversarial examples and counterfactual explanations have largely been growing independently. 
This has led to several recent works trying to elucidate their similarities and differences. 
Most prominently, it has been argued that adversarial examples, as opposed to counterfactual explanations, have a unique characteristic in that they lead to a misclassification compared to the ground truth. 
However, the computational goals and methodologies employed in existing counterfactual explanation and adversarial example generation methods often lack alignment with this requirement.
Using formal definitions of adversarial examples and counterfactual explanations, we introduce non-adversarial algorithmic recourse and outline why in high-stakes situations, it is imperative to obtain counterfactual explanations that do not exhibit adversarial characteristics. We subsequently investigate how different components in the objective functions, e.g., the machine learning model or cost function used to measure distance, determine whether the outcome can be considered an adversarial example or not. 
Our experiments on common datasets highlight that these design choices are often more critical in deciding whether recourse is non-adversarial than whether recourse or attack algorithms are used.
Furthermore, we show that choosing a robust and accurate machine learning model results in less adversarial recourse desired in practice.

\end{abstract}
\setcounter{footnote}{0}
\section{Introduction}
A continuous stream of predominantly independent research in the fields of adversarial examples \citep{szegedy2013intriguing,goodfellow2014explaining} and counterfactual explanations \citep{wachter2017counterfactual,Ustun2019ActionableRI,pawelczyk2019,verma2020counterfactual} has sparked an ongoing scholarly discourse on their similarities and differences \citep{pawelczyk2022exploring,freiesleben2022intriguing}.
While adversarial examples originate from the security literature, characterizing instances capable of deceiving machine-learned classifiers into erroneous decisions, algorithmic recourse has its roots in the trustworthy machine-learning literature. Algorithmic recourse is primarily concerned with providing actionable recommendations for changes that would lead to a different, more favorable outcome for the end user (e.g., changing a loan decision from rejection to acceptance). Despite the apparent differences in goals and associated semantics between adversarial examples and recourse, scholars have observed a strikingly similar algorithmic foundation underpinning these two domains \citep{freiesleben2022intriguing,pawelczyk2022exploring}.

\begin{figure*}[t]
\centering
\input{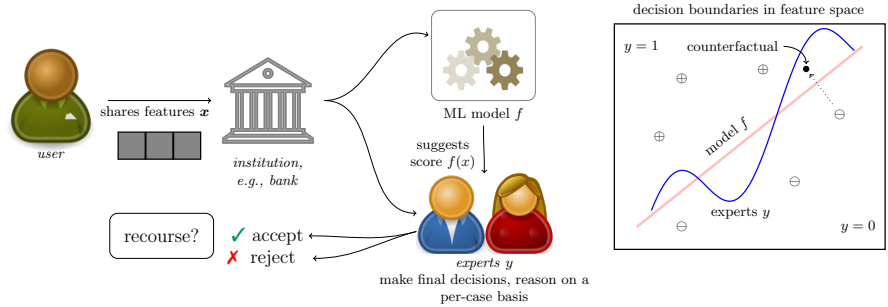}
\caption{
\textbf{Overview of the realistic decision-making scenario considered in this work.} We consider the relevant case where an institution, e.g., a bank, deploys a machine learning model to support decision-making overseen by human experts that make final, case-based decisions based on the model's score (left). In such a setting, constructing recourse only based on the scoring model $f$ may lead to unreliable recourse because the experts' final $y$ decision is based on further restrictions, thereby representing a shifted decision boundary (right).\label{fig:teaser_nadv}}
\end{figure*}

The current debate surrounding the potential distinctions between these two concepts remains somewhat ambiguous. 
To provide greater context and significance to this discourse, we establish a tangible connection to a real-world application where the differentiation between counterfactual and adversarial examples becomes intuitive and indispensable. To this end, we slightly modify the established recourse problem in the context of loan assignments \citep{Ustun_2019}. 
Unlike previous work, which assumes that a machine learning system solely determines loan assignments, we argue that this perspective oversimplifies the real world. Article 22 of the European Union's General Data Protection Regulation (GDPR) \citep{regulation2016regulation}, which asserts the right of the data subject \emph{``not to be subject to a decision based solely on automated processing which produces legal effects concerning him or her''}, thereby suggesting that automated models alone cannot make legally binding decisions. Consequently, we consider a more practical scenario where algorithmic decisions are subject to scrutiny by a human expert panel. This expert panel holds the authority to issue a final, case-specific decision and can override the model's recommendation. 
This refined setup is illustrated in \Cref{fig:teaser_nadv}.


Complementarily, the GDPR grants individuals who receive an adverse decision the right to receive \emph{``meaningful information about the logic involved''} which, in a broader context, can be interpreted as the right to ``recourse'' \citep{voigt2017eu}. When the model exclusively determines decisions, it is evident that recourse can be directly computed from the model itself. However, in the more realistic scenario considered in this work, where human experts play a role in the final decisions, the model's output does not fully encapsulate the ultimate decision. This raises the question of appropriate recourse design in such a scenario and how to reconcile these two GDPR principles -- the right to receive meaningful recourse and the prohibition of fully automated decision-making.

Under the premise that the model has been mainly distilled from past decisions of the experts, we consider the experts as an imperfect oracle providing ground truth labels,\footnote{The oracle is imperfect as some labels are generated from ``defaults'', i.e., false positives of expert decisions.} whereas the model returns an imperfect approximation of these labels. While the humans decide on a per-case basis, it is hard to directly ask them for specific thresholds, as the interplay  of the features quickly makes the task intractable.
Therefore, we are interested in computing counterfactual explanations that do not only change the model's prediction but also flip the \emph{true} labels. This perspective aligns with the argument made by Freiesleben~\cite{freiesleben2022intriguing} that a distinctive feature of adversarial examples, as opposed to counterfactual explanations, is their tendency to be misclassified regarding their true labels. Since counterfactual explanations should also change the true label in this case, this gives rise to the term ``non-adversarial algorithmic recourse", i.e., \emph{counterfactual explanations that come with both a change in the model's prediction and a changed ground-truth label}.

Unlike prior work taking a merely definitional view, this work additionally contributes to implementing non-adversarial algorithmic recourse in practical scenarios. In summary, we propose the following contributions:
\begin{itemize}
    \item \textbf{Introduction of a novel recourse problem:} We introduce a novel recourse problem that addresses real-world decision systems wherein human experts play a pivotal role in making case-based decisions, while  also considering input from a machine learning model. 
    \item \textbf{Proposing non-adversarial recourse as a solution to the realistic recourse problem:} We consider prior work's \cite{freiesleben2022intriguing} distinction of adversarial examples and counterfactual explanations and suggest a novel formal definition of \emph{non-adversarial algorithmic recourse}, proving a conceptual bridge between the academic discourse on distinguishing adversarial examples from counterfactual explanations and practical decision-making.
    \item \textbf{Promoting non-adversarial recourse theoretically:} After a theoretical analysis of the problem, we derive optimal cost functions that encourage non-adversarial recourse. Our analysis underscores how feature attributions can be leveraged to identify task-relevant features contributing to less ``adversarial'' recourse. 
    \item \textbf{Empirical Insights:} We are the first to consider several other key components practitioners can manipulate to foster non-adversarial algorithmic recourse. These include improving the robustness and accuracy of the machine learning model and the recourse algorithms. In contrast to parts of the literature which argue that loss functions are central, we empirically find that changes in the model are often more significant than the cost function.
\end{itemize}

\section{Related Work}\label{sec:related_work}

\textbf{Human-Assisted Decisions.} In crucial situations, societies rely on human experts for decisions. However, delays and quality issues due to a shortage of experts and a high volume of decisions, e.g., long waits for medical diagnoses, have sparked a debate on when automated or human decision-making should be deployed. A stream of prior works \cite{topol2019high,cheng2015antisocial,pradel2018deepbugs} argue that ML models should make decisions in high-stake domains where they have matched or surpassed the average of human performance. Nevertheless, their decisions can still be worse than those of human experts \cite{raghu2019algorithmic} in some cases. In this direction, works such as \cite{de2020regression,pmlr-v119-mozannar20b,de2021classification} propose to optimize ML models to operate under different automation levels: i.e., take decisions on a fraction of the given instances and leave the rest to human experts.
In line with other works \cite{DBLP:conf/iui/FerreiraM21}, we argue that the human factor in the loop in a human-AI partnership cannot be neglected when considering the application of AI on real-world problems \cite{Grudin_2009,abdul2018trends}. This position is also cemented in common data protection laws such as the EU's GDPR \cite{regulation2016regulation}, which grants individuals a right to object fully automated decision-making. For GDPR-compliant decision-making, human oversight can thus be considered essential. Unlike previous works, we explicitly model a human expert panel in the decision-making setup as depicted in \Cref{fig:teaser_nadv}, which makes the generation of reliable recourse much more challenging. 

\textbf{Counterfactual Explanations.} There is an established literature on the computation of counterfactual explanations \cite{wachter2017counterfactual,mothilal2020explaining,ma2022clear,carreira2021counterfactual,poyiadzi2020face,abrate2021counterfactual,rawal2020beyond,Ustun2019ActionableRI,konig2023improvement} in variegated domains. According to Guidotti et al.~\cite{guidotti2022counterfactual}, given a classifier $f$ that outputs a decision $f(\vx) = y$ for an instance $\vx$, a counterfactual explanation of $\vx^\prime$ is an instance $\vx^\prime$ such that $f(\vx^\prime
) \neq y$, and the difference between $\vx$ and $\vx'$ is minimal. Current research streams include the robustness of counterfactual explanations \cite{upadhyay2021robust,dominguez-olmedo22a,pawelczyk2023probabilistically} and the compatibility with other GPDR principles \cite{pawelczyk2023on}. We briefly review this research field in the following but point the reader to recent surveys \cite{verma2020counterfactual,guidotti2022counterfactual,prado2023survey} for a comprehensive overview.
Mothilal et al. \cite{mothilal2020explaining} solves an optimization problem with various constraints, among which user-specified ones for (im)mutable features, to ensure feasibility and diversity when producing a set of counterfactuals for a given input. Carreira-Perpi{\~n}{\'a}n and Hada \cite{carreira2021counterfactual} propose CEODT specifically designed for classification trees, including Oblique Decision Trees (ODTs) \cite{heath1993induction}. Because the counterfactual optimization problem for ODTs is non-convex, nonlinear, and non-differentiable, CEODT computes an exact solution via the optimization problem within the region represented by each leaf and finally picks the leaf with the best solution. 
Lastly, Ustun et al. \cite{Ustun2019ActionableRI} were among the first authors to address the problem of actionability in counterfactual explanations (i.e., recourse). Their method constrains the generated counterfactuals such that manipulations do not change immutable features. Overall, we note that previous literature relies on the common assumption that an automated model acts as a sole decision-maker, which might not be realistic in practical scenarios.

\textbf{Adversarial Examples.} Following Szegedy et al.~\cite{szegedy2013intriguing}, \advexs are instances that contain subtle perturbations -- usually via adding small amounts of noise -- to instances in the training set. These ``new'' instances, when fed to an underlying ML model, produce an erroneous output with high confidence. 
It is possible to build an \advex $\vx'$ which is indistinguishable\footnote{We invite the reader to think about images in this context, as described in \cite{goodfellow2014explaining}. Additionally, some works that analyze perturbations - e.g., adversarial patches  - that are perceptually distinguishable by humans but fool the classifier $f$ \cite{demir2018patches,zhao2022ap,Du_2022_WACV}.} from $\vx$ but is classified incorrectly, i.e., $f(\vx') \neq y$. Successfully generating such examples gives rise to \textit{adversarial attacks}~\citep{goodfellow2014explaining,moosavi2017universal,baluja2017adversarial}, which can have potentially lethal consequences (e.g., in biosecurity and biotechnology \cite{pauwels2023protect}, autonomous driving \cite{Duan_2021_CVPR,zhang2022evaluating}, and power grid blackouts \cite{garcia2017hey}). Several methods have been proposed in the literature to generate \advexs assuming varying degrees of knowledge/access of the model, training data, and methods for injecting perturbations. Goodfellow et al.~\citet{goodfellow2014explaining}, Kurakin et al. \cite{kurakin2016adversarial}, and Moosavi et al., \citet{moosavi2016deepfool} propose methods with gradient and data access to find the minimum $\ell_{\infty}$-norm (and $\ell_2$-norm respectively) perturbations.
With only assuming query access to the target classifier, the authors in \citet{su2019one,croce2019sparse,narodytska2016simple} design \advexs to tightly control sparsity. 
We refer the reader to a well-established survey for a comprehensive overview of \advexs \citep{akhtar2018threat}. 

\textbf{Linking Counterfactuals and Adversarial Examples.} Strikingly, \counterexps and \advexs have conceptual connections and a similar formulation \cite{freiesleben2022intriguing,browne2020semantics,wachter2017counterfactual} (see also Sec. \ref{sec:preliminaries}).
Freiesleben~\citet{freiesleben2020counterfactual} highlights conceptual differences in aims, formulation, and use-cases between both sub-fields and suggests that the distinctive formal feature of \advexs lies in their misclassification concerning the ground truth. 
Concurrently, there have been proposals to align recourse with a ground truth. König et al.~\cite{konig2023improvement} proposes improvement-focused causal recourse (ICR), designed to change the true targets instead of the predictions but relies on causal information. Laugel et al.~\cite{laugel2019dangers} proposes the notion of ``justified recourse'' that should be close to a correctly classified instance. On the other hand, Browne et al. \citet{browne2020semantics} focus on deep networks and attribute conceptual differences to the interpretation of semantics in the hidden layers of deep networks. Pawelczyk et al.~\cite{pawelczyk2022exploring} formalize the similarities between popular \counterexps and \advex generation methods identifying conditions when they are equivalent.
Trying to disentangle and reconcile the various distinctions made in prior works, we provide formal definitions in the next section. Besides König et al.~\cite{konig2023improvement}, which relies on causal information, there have been few attempts to implenent recourse that follows the ground truth. In this work, we valuable insights on how to implement non-adversarial recourse in practical decision-making. 


\section{Preliminaries}\label{sec:preliminaries}
We first formalize the general problem considered in this work, before we provide the relevant distinctions between adversarial examples and counterfactual explanations.

\subsection{The general problem}
Both recourse and adversarial methods consider a fixed machine learning model $f: \mathcal{X} \rightarrow \mathcal{Y}$. We usually consider the binary classification problem, where the label is binary, i.e.,  $\mathcal{Y} = \left\{0,1\right\}$ or a numerical score, $\mathcal{Y} = \mathbb{R}$.

We suppose there is another function $y:\mathcal{X} \rightarrow \mathcal{Y}$ that assigns the true labels and represents the human experts in our introductory example. In practice, it is impossible to perfectly learn this function with a model, for instance due to insufficient data coverage or additional circumstances that can be taken into considerations only by the human experts. However, it is possible query $y$ a limited number of times, as it is possible to present the experts with an example and ask for their decision. We model the expert predictions $y$ in the scenario outlined as 
\begin{align}
    y(\vx) = g(\vx, f(\vx)),
\end{align}
where $g$ models the human expert committee that can recalibrate the score in light of the features in their entirety. However, we suppose that we usually have $y(\vx) \approx f(\vx)$, i.e., the original score is only lightly adapted through $g$.

As noted before \cite{pawelczyk2022exploring}, the classical optimization problem solved by both practical adversarial and counterfactual methods for a model $f: \mathcal{X} \subset \mathbb{R}^{k}\rightarrow \mathcal{Y}\subset \mathbb{R}$, a factual input $\vx \in \mathcal{X}$, and a target label $y_t \in \mathcal{Y}$ is mathematically similar and can usually be formalized as a special case of the following general optimization problem \cite{freiesleben2022intriguing}:
\begin{equation}\label{eqn:recourseproblem}
    \underset{\vx^{'} \in \mathcal{X}}{\text{\normalfont argmin}}~d_1(\vx,\vx^{\prime})+\lambda d_2(f(\vx^{\prime}),y_t),
\end{equation}
where $d_1: \mathcal{X} \times \mathcal{X} \rightarrow \mathbb{R} $ is a distance metric defined on the input space, $d_2: \mathcal{Y} \times \mathcal{Y} \rightarrow \mathbb{R}$ is a metric on the output space and $\lambda \in\mathbb{R}_{\geq 0}$ is a non-negative trade-off parameter. Intutively, the solution to this problem returns instances, that are close to the factual $\vx$ and have a label that is close (or corresponds exactly) to the target label $y_t$.

\subsection{Algorithms for Computing Counterfactual Explanations and Adversarial Examples}
We briefly review the most common strategies to compute counterfactuals and adversarial examples in practice.
\paragraph{Score CounterFactual Explanations (SCFE).}
For a given classifier $f(h(\vx))$ that relies on logit scores $h(\vx)$ and a distance function $d: \mathcal{X} \times \mathcal{X} \to \mathbb{R}_{+}$, Wachter et al. \citet{wachter2017counterfactual} formulate the problem of finding a \counter $\vx'$ for $\vx$ as:
\begin{equation}
{\text{\normalfont argmin}}_{\vx'} ~ \big(h(\vx')-s\big)^{2} + \lambda \, d(\vx,\vx'),
\label{eq:scfe_objective}
\end{equation}
where $s$ is the target score for $\vx$. The problem is solved for different values of $\lambda$ until $f(\vx') = s$. 
More specifically, to arrive at a counterfactual probability of 0.5, the target score for $h(\vx)$ for a sigmoid function is $s=0$. Using the inverse logit transform $h(\vx)=\text{invlogit}(f(\vx))$, the first part of the objective can be interpreted as a particular instantiation of $d_2$ in Eqn.~\eqref{eqn:recourseproblem} when $\mathcal{Y}$ is taken to be the interval $\lbrack 0,1\rbrack$.

\paragraph{Diverse Counterfactual Explanations (DiCE).} As different users may have different preferences (i.e., it might be easier for them to change one feature or another), DiCE \cite{mothilal2020fat} generates multiple counterfactuals. An additional loss term is added to the objective in \Cref{eq:scfe_objective} to encourage diversity. As users will only choose one counterfactual in practice, we usually consider a randomly selected instance of the discovered recourse candidate for evaluation as in \cite{pawelczyk2023on}.

\paragraph{Actionable Recourse (AR).} The actionable recourse (AR) method by Ustun et al.~\cite{Ustun2019ActionableRI} sets up the following optimization problem:
\begin{align}
    \min~ &\text{cost}(\bm{\delta}; \vx)\\
&\text{s.t.} f (\vx +\bm{\delta}) = +1,
\delta \in \mathcal{A}(\vx),
\end{align}
where $+1$ corresponds to the the positive outcome and a is an action set $\mathcal{A}(\vx)$. This problem corresponds to Eqn.~\eqref{eqn:recourseproblem} when using a distance function $d_1$ that returns $\infty$ once $\delta \neq \mathcal{A}(\vx)$ and the cost function otherwise. The distance $d_2$ can be interpreted as the Dirac-distance, that is $\infty$ once $ f (\vx +\bm{\delta}) \neq 1$. They solve the problem using mixed integer linear programming (MIP) for linear models.

Like counterfactual explanations, most adversarial example methods also solve a constrained optimization problem to find perturbations in the input manifold that cause models to misclassify.

\paragraph{C\&W Attack.}
For a given input $\vx$ and classifier $f$, Carlini and Wagner ~\citet{carlini2017towards} formulate the problem of finding an adversarial example $\vx' = \vx{+} \bdelta$ such that $f(\vx') \neq f(\vx)$ as:
\begin{equation}
\underset{\vx^{'} \in \mathcal{X}}{\text{\normalfont argmin}}
\;c \cdot \ell(\vx')+d(\vx, \vx')~~\mbox{s.t.}~~\vx' \in [0,1]^{d},
\label{eq:cw_objective}
\end{equation}
where $c > 0$ is a suitably chosen hyperparameter, and $\ell(\cdot)$ is an objective function on the adversarial $\vx'$ s.t. $f(\vx') = y_t$ iff $\ell(\vx') \leq 0$ with $y_t$ being a target class. The authors choose $d(\vx,\vx')$ to be the $l_p$ norm of $\bdelta$, i.e., minimizing the $p$-norm of $\bdelta$ is equivalent to minimizing $d(\vx,\vx')$.

\paragraph{DeepFool Attack.}
For a given instance $\vx$, DeepFool~\citep{moosavi2016deepfool} perturbs it by adding small perturbation $\bdelta_{\text{DF}}$ at each iteration. 
The minimal perturbation to change the classification model's prediction is the solution to the following objective:
\begin{align}
\begin{split}
\bdelta^{*}_{\text{DF}}(\mathbf{x}) \in \underset{\bdelta { \text{ s.t. } } \vx + \bdelta \in \mathcal{X}}{\text{\normalfont argmin}} ||\bdelta||_{2} 
\text{ s.t. } \text{sign}(f(\mathbf{x} + \bdelta)) \neq \text{sign}(f(\mathbf{x}))
\end{split}
\label{eq:deepfool_objective}
\end{align}

\paragraph{PGD Attack.} PGD \cite{madry2017towards} is a first-order optimization technique. In the context of adversarial examples, it is usually used to maximize\footnote{Thus, projected gradient ascent is often the more appropriate description for this attack. However, we will follow common practice and refer to the algorithm as PGD.}, the objective for a specific factual $\vx$. This is because the objective is typically chosen to be the cross-entropy loss $\mathcal{L}$:
 \begin{equation}\label{eq:pgd_max}
    \underset{\bdelta \text{ s.t. } \vx + \bdelta \in \mathcal{C}}{\text{\normalfont argmax}} \; \mathcal{L}(f(\vx + \bdelta), f(\vx))
 \end{equation}
where $\bdelta$ is the adversarial perturbation to be added to the factual $\vx$. PGD maximizes the objective by taking steps along the gradient's direction. 
After each update, the current perturbation $\bdelta^t$ is projected onto a set of constraints $\mathcal{C}$. 
For instance, the adversarial examples are all constrained to a ball of size $\epsilon$ around $\vx$. 
We argue that the projection of the adversarials $\vx' = \vx + \bdelta$ into an $\epsilon$-ball could be interpreted as a $d_1$ distance function in Eqn.~\eqref{eqn:recourseproblem}, that returns an infinite cost value for actions outside the $\epsilon$-ball. Meanwhile, the cross-entropy loss subsumes the role of the $d_2$ function. Therefore, Eqn.~\eqref{eq:pgd_max} can be considered as a special case of  Eqn.~\eqref{eqn:recourseproblem} transformed into a maximization problem.

We invite the reader to notice that the approaches presented above -- whether adversarial attacks or counterfactual explanation methods -- solve the same objective. In fact, they can be interpreted as heuristics to optimizing an instance of the formulation in  \Cref{eqn:recourseproblem}, although pertaining to different ``semantics'' as argued in \cite{sahil2010counterfactual}. However, a precise distinction between counterfactual explanations and adversarial attack algorithms cannot be derived from their implementations. To this end, we investigate precise definitions for both problems in the next section.


\section{Definitions}
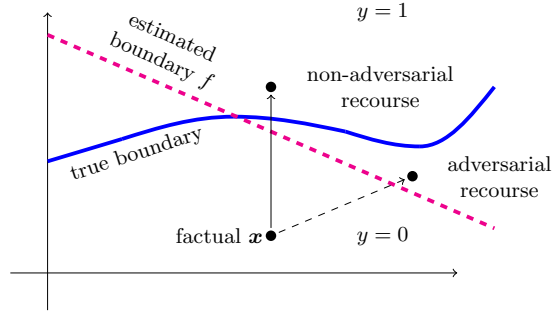
\begin{figure}[t]
\centering
\scalebox{0.9}{
\begin{tikzpicture}[scale=1.1]
\draw[->] (-1.5,0) -- (4.5,0) node[right] { };
\draw[->] (-1,-0.5) -- (-1,3.5) node[above] { };
\node at (3.5, 3.5) {$y=1$};
\node at (3.5, 0.5) {$y=0$};
\node[rotate= 18] at (0.2,1.6) {true boundary};
\draw[ultra thick, blue] (-1,1.5) -- (0,1.8);
\draw[ultra thick, blue] (0,1.8) sin (1.5,2.1);
\draw[ultra thick, blue] (1.5,2.1) cos (3,1.9);
\draw[ultra thick, blue] (3,1.9) sin (4,1.7);
\draw[ultra thick, blue] (4,1.7) cos (5,2.5);
\draw[ultra thick, dashed, magenta] (-1,3.2) -- (5,0.6);
\node[rotate=-24] at (0.6,3.0) {\parbox{3cm}{\centering estimated\\boundary $f$}};
\fill[black] (2,0.5) circle (2pt) node (factual) {};
\node[left] at (2,0.5) {factual $\vx$};

\fill[black] (2,2.5) circle (2pt) node (nonadvcf) {};
\node[right] at (2,2.5) {\parbox{3cm}{\centering non-adversarial\\ recourse}};

\node[right] at (3.6,1.3) {\parbox{3cm}{\centering adversarial\\ recourse}};
\fill[black] (3.9,1.3) circle (2pt) node (advcf) {};

\draw[->, dashed] (factual) to (advcf);
\draw[->] (factual) to (nonadvcf);
\end{tikzpicture}}
    \caption{\textbf{Visualizing our definitions.} The space of valid recourse for a factual $\vx$ changes crosses the classifier $f$'s estimated decision-boundary (pink). The experts combine it with their expertise and restrictions into a latent decision boundary  (blue). However, some recourse might not change the true label and is therefore considered adversarial (dashed arrow). The challenge is to obtain recourse that convinces the human experts. To this end, we are interested in finding the directions that lead to \textit{non-adversarial recourse} (solid arrow).\label{fig:vizdefs}} 
\end{figure}

\subsection{Formalizing Adversarials and Counterfactuals}
We take the definition of an adversarial example by Freiesleben \cite{freiesleben2022intriguing} as a starting point. It intuitively describes the properties that such instances should have. In other words, they should be close to the original instance, change the model's predictions and be misclassified. Most notably and in contrast to other works, Freiesleben argues that the misclassification is a distinctive property of adversarial examples. This distinctive property has also previously been mentioned in other works on adversarial examples more or less directly \cite{stutz2019disentangling}, giving rise to the following definition:

\begin{definition}[Adversarial Example \cite{freiesleben2022intriguing}] An instance $\vx^\prime  \in \mathcal{X}$ is an \textbf{adversarial example} for a factual $\vx \in \mathcal{X}$ and a classifier $\vf: \mathcal{X} \rightarrow \mathcal{Y}$ if the following conditions hold:
\begin{enumerate}
\item[(1)] $\vx^\prime$ is close to $\vx$, i.e., $d_1(\vx, \vx^\prime) < \epsilon$;
\item[(2)] the classifier output is changed, i.e., $f(\vx) \neq f(\vx^\prime)$;
\item[(3)] $\vx^\prime$ is misclassified, i.e., $y(\vx^\prime) \neq f(\vx^\prime)$.
\end{enumerate}
\end{definition}
We also consider the definition of recourse (or equivalently, counterfactual examples) by Freiesleben \cite{freiesleben2022intriguing}, which states that recourse $\vx'$ changes the classification label and is the closest point to the factual that does so. We propose a slight relaxation. In particular, we argue that even points that are not closest to the factual are still valid (though possibly suboptimal) recourse. 
\begin{definition}[Recourse] An instance $\vx^\prime  \in \mathcal{X}$ is a \textbf{recourse} for a factual $\vx \in \mathcal{X}$ with $f(\vx) \neq y_t$, a classifier $\vf: \mathcal{X} \rightarrow \mathcal{Y}$, and a target label $y_t \in \mathcal{Y}$ if the following conditions hold:
\begin{enumerate}
\item[(1)] $\vx^\prime$ is close to $\vx$, i.e., $d_1(\vx, \vx^\prime) < \epsilon$;
\item[(2)] the classifier output is changed, i.e., $f(\vx) \neq f(\vx^\prime)$.
\end{enumerate}
\end{definition}

These general definitions cover most definitions explicitly or implicitly used in the literature (see \cite{freiesleben2022intriguing} for details). We immediately see that our definition of recourse abandons the final constraint in the definition of adversarial examples. This implies that, according to these definitions, (a) all adversarial examples are recourse, and (b) there is a distinct (though potentially empty) subset of examples, that are recourse, but are not adversarials, as visualized in \Cref{fig:vizdefs}. 

\subsection{Non-Adversarial Algorithmic Recourse}

In this work, we place our attention on the examples present in this subset, that are recourse but not adversarial examples. We thus refer to them as \emph{non-adversarial recourse} and introduce a novel definition for this class of instances:

\begin{definition}[Non-Adversarial Recourse] An instance $\vx^\prime  \in \mathcal{X}$ is \textbf{non-adversarial recourse} for a factual $\vx \in \mathcal{X}$ and a classifier $\vf: \mathcal{X} \rightarrow \mathcal{Y}$ if the following conditions hold:
\begin{enumerate}
\item[(1)] $\vx^\prime$ is close to $\vx$, i.e., $d_1(\vx, \vx^\prime) < \epsilon$;
\item[(2)] the classifier output is changed, i.e., $f(\vx) \neq f(\vx^\prime)$;
\item[(3)] $\vx$ is not misclassified, $f(\vx^\prime)=y(\vx^\prime)$.
\end{enumerate}
\end{definition}

We observe that in the considered realistic decision-making scenario, we desire recourse that convinces the human experts, i.e., also changes the true label $y$. These correspond exactly to the instances described in the definition of non-adversarial recourse. 



\section{Theoretical Analysis}\label{sec:theory}
As outlined in \Cref{fig:vizdefs}, we are interested in finding changes, or at least directions of change, that lead to non-adversarial recourse efficiently. As it is impossible to precisely model the ground truth $y$ in our setup (otherwise, there would be no need for an additional human expert), this is  challenging in practice. However, we can use some guidance from the model, which approximates the ground truth, to find non-adversarial recourse. 

\subsection{Summarizing influential factors for less adversarial recourse}
We first take a step back and consider the general formulation of the problem given in Eqn.~\eqref{eqn:recourseproblem}. We observe that the problem formulation features three potential factors of influence (the model $f$, the distance functions $d_1$ and $d_2$) and a hyperparameter (the choice of optimization algorithm) that can be changed in practice to arrive at less adversarial recourse. If we follow the usual binary classification setup where we chose $\lambda > 0$ and $d_2$ to be the Dirac distance that amounts to infinity if the target label is not met, i.e., $d_2(f(\vx^{\prime}),y_t) = \delta_{f(\vx^\prime)=y_t}$, there are three remaining factors of influence, that we tackle in this study with different outcomes:

\paragraph{Machine learning model.} Considering the model $f$ first, we note that there is a simple theoretical solution to non-adversarial recourse:  If the model would exactly match the theoretical ground truth, i.e., $f \equiv y$, there would be no adversarial recourse as every instance that leads to a different model prediction also changes the ground truth. However, in the setup we consider, it is impossible to perfectly learn $y$. Nevertheless, using the best possible model as close to the ground truth as possible should be fruitful. 
Another way to improve the model's alignment with the ground truth -- in case the truth is known to be smooth in some measure -- could be to potentially leverage regularization techniques such as adversarial training \cite{madry2017towards} to rule out many adversarial instances in the first place.
\textbf{We empirically find that more accurate and robust models lead to less adversarial recourse.}

\paragraph{Input space distance function.} The distance function $d_1$ has been attributed a crucial role when computing recourse or adversarial examples. For instance, Wachter et al. \cite{wachter2017counterfactual} have claimed that, unlike recourse, \emph{none of the standard works on adversarial perturbations use appropriate distance functions}. In this work, we follow the perspective of \cite{browne2020semantics,freiesleben2022intriguing}, who argue that the distance metric is not definitional but may still play an essential role in making recourse non-adversarial. Besides standard cost functions like $p$-norms such as the $l_1$, $l_2$, and $l_\infty$, we are interested in how feature weightings may potentially impact recourse. We follow the intuition that some features are discriminative in the ground truth problem, e.g., income determines creditworthiness. However, ML models may rely on many more features, as the model designers cannot precisely specify a priori which features will be relevant for the task. When non-discriminative features are used in the task, they may open the door to adversarial changes as they can be picked up by an ML model regardless of their irrelevance w.r.t. the ground truth. In the next section, we will present an attempt to down-weigh the influence of such features by individually assigning a cost to each of them. In particular, we will consider distance functions of the form 
\begin{align}
    d_{1, S}(\vx, \vx^\prime) \coloneqq \vdelta^\top \mS \vdelta, \mS = \text{diag}\left(\vs \right), \label{eqn:distancefn}
\end{align}
where $\mS \in \mathbb{R}^{k\times k}$ is some diagonal matrix with diagonal $\vs = [s_1, \ldots, s_k]^\top \in \mathbb{R}_{>0}^k$ and $\vdelta \coloneqq \vx^\prime -\vx$. 
We will see that the problem of finding optimal values for $\vs$ can be solved analytically based on the gradients for linear models. \textbf{Surprisingly, we empricially find that the cost function does not play a key role in obtaining non-adversarial recourse.}

\paragraph{Optimization routine} As the general problem is highly non-linear for complex models, it is hard to discover an optimal solution. As a result, algorithms to compute recourse or adversarial examples include different heuristic optimization routines such as stochastic gradient descent (deployed in SCFE, DICE, and C\&W), gradient projection (deployed in PGD), or discretization (deployed in AR). The optimization procedure may thus also play a non-negligible role in determining whether the nature of the resulting recourse is adversarial and whether approaches designed for recourse yield fewer adversarial examples than their adversarial counterparts. \textbf{In this regard, we find that adversarial methods succeed to compute non-adversarial recourse, but also incur higher costs.}

\subsection{Optimal cost functions under linear models with noisy labels}
\label{sec:optimalcostfunctions}
In this section, we will restrict ourselves to the input space distance function $d_1$ and study its influence on the recourse from a theoretical standpoint.

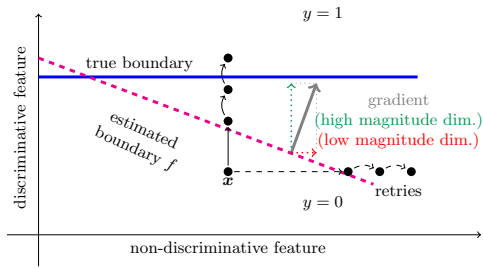
\begin{figure}[t]
\centering
\scalebox{0.7}{
\begin{tikzpicture}[scale=1.2]
\draw[->] (-1.5,0) -- (6,0);
\node at (2,-0.2) {non-discriminative feature};
\draw[->] (-1,-0.5) -- (-1,3.5);
\node[rotate=90] at (-1.3,1.7) {discriminative feature};
\node at (3.5, 3.5) {$y=1$};
\node at (3.5, 0.5) {$y=0$};
\node at (0.6, 2.7) {true boundary};
\draw[ultra thick, blue] (-1,2.5) -- (5 ,2.5);
\draw[ultra thick, dashed, magenta] (-1,2.8) -- (4.3,0.8);
\draw[->, ultra thick, gray] (3,1.3) -- (3.4,2.4);
\draw[->, ForestGreen, thick, dotted] (3,1.3) -- (3,2.4);
\draw[->, red, thick, dotted] (3,1.3) -- (3.4,1.3);
\draw[gray, dotted] (3.4,1.3) -- (3.4,2.4);
\draw[gray, dotted] (3,2.4) -- (3.4,2.4);
\node at (4.7, 1.8) {\parbox{5cm}{\centering \textcolor{gray}{gradient}\\ \footnotesize{\textcolor{ForestGreen}{(high magnitude dim.)}} \\
\textcolor{red}{(low magnitude dim.)}}};
\node[rotate=-24] at (0.6,1.5) {\parbox{3cm}{\centering estimated\\boundary $f$}};
\node[below] at (2,1) {$\vx$};
\fill[black] (2,1) circle (2pt) node (factualr) {}; 
\fill[black] (2,2.8) circle (2pt)  node (retrya2) {};
\fill[black] (2,2.3) circle (2pt) node (retrya1) {};
\fill[black] (2,1.8) circle (2pt) node (nonadvcf) {};

\fill[black] (3.9,1) circle (2pt) node (advcf) {};
\fill[black] (4.4,1) circle (2pt) node (retryb1) {};
\fill[black] (4.9,1) circle (2pt) node (retryb2) {};

\draw[->, dashed] (factualr) to (advcf);
\draw[->] (nonadvcf) to[out=120, in=240] (retrya1);
\draw[->] (retrya1) to[out=120, in=240] (retrya2);
\draw[->] (factualr) to (nonadvcf);
\draw[->, dashed] (advcf) to[out=30, in=150] (retryb1);
\draw[->, dashed] (retryb1) to[out=30, in=150] (retryb2);

\node at (4.7, 0.7) {retries};
\end{tikzpicture}}
    \caption{\textbf{Role of discriminative features in providing non-adversarial recourse.} When features can be discriminative, (i.e., class-relevant) or non-discriminative (i.e., noise features), exploiting the discriminative ones will eventually lead to non-adversarial recourse, whereas solely relying on the non-discriminative ones will result in an adversarial. Nevertheless, even when selecting the correct features, several retry steps in the recourse direction may be required to cross the true decision boundary. To align recourse with discriminative features, the gradients of the model may serve as guidance, as we expect the discriminative dimensions to exhibit a \textcolor{ForestGreen}{higher} gradient magnitude. \label{fig:discfeatureretry}} 
\end{figure}

We first introduce a measure to quantify the extent to which recourse is non-adversarial. To be able to do so, we consider the simplified setup where we have a feature set $\mathcal{F}$ and a subset of discriminative features $\mathcal{F}_{\text{disc}} \subset \mathcal{F}$ that contains relevant information affecting the ground truth. The remainder of the features are noise variables. Such features exist for many tasks; however, they may not be axis-aligned initially. For instance, in generative image generation models such as StyleGAN \cite{karras2020analyzing}, the first latent variables control high-level concepts in the generation, whereas the later variables merely add noise that is unimportant for the classification output. Successes with dimensionality reduction techniques through autoencoding \cite{ilkhechi2020deepsqueeze} also show that important information occupies only a subspace of tabular data. As outlined in \Cref{fig:discfeatureretry}, following the discriminative features is essential for obtaining non-adversarial recourse. We can quantify the share of the recourse that lies in the discriminative directions over the entire length of the recourse vector through the following measure.
\begin{definition}[NADV measure]
Let $p \in \mathbb{N} \cup \left\{\infty \right\}$. The non-adversarialness measure $\text{NADV}_p$ is defined as
\begin{align}
    \text{NADV}_p(\vdelta) = \frac{\sum_{i\in \mathcal{F}_{\text{disc}}} \lvert\vdelta_i\rvert}{\lVert\vdelta\rVert_p}.
\end{align}
\end{definition}

We consider linear models in our initial analysis, as they are the standard in many industries (e.g., finance, healthcare, real estate, e-commerce, and marketing) and are commonly studied in the literature on algorithmic recourse \cite{pawelczyk2023on,upadhyay2021robust}. They model a generative process of the form
\begin{align}
y(\vx) = \vbeta^\top\vx + \epsilon,\label{eq:linearprocess}
\end{align}
where $\epsilon \sim \mathcal{N}(0, \sigma^2)$ is Gaussian noise of variance $\sigma^2$ and $\beta \in \mathbb{R}^k$ denotes the true linear parameter vector. Such a model can be easily adapted to a classification task by introducing a decision threshold, e.g., $y(\vx) > 0$ indicates a positive outcome. As motivated in the introduction, the noise may represent uncertainty and variance in the human labels. We are interested in weightings $s_i$ that minimize this measure, potentially leveraging the empirical coefficients $\bm{\hat\beta}$ obtained when fitting a linear model to the noisy data. 

\begin{theorem}[Optimal feature weights for recourse in linear models]\label{thm:recourseweight} Suppose the data-generating process in Eqn.~\eqref{eq:linearprocess} and that for $i \notin \mathcal{F}_{\text{disc}}$, we have $\beta_i=0$, and for $i \in \mathcal{F}_{\text{disc}}$, $|\beta_i| > \alpha \in \mathbb{R}$. We can maximize the expected NADV$_p$ measure for $p\in \{1,2,\infty\}$ when using the empirical coefficients $\hat{\beta}_i$ of the fitted model by setting the weights to
\[
s_i\sim \begin{cases*}
     \left\{1, \text{~if~}i{=}\argmax_j \vp_{\text{disc}}(\hat{\beta}_j), \text{~else~} \infty \right\} & if $p=1$ \\
     \frac{\lvert\hat{\beta}_i\rvert}{\vp_{\text{disc}}(\hat{\beta}_i)} & if $p=2$ \\
    \lvert \hat{\beta}_i \rvert & if $p=\infty$
    \end{cases*},
\]
where $\vp_{\text{disc}}(\hat{\beta}_i)$ is a probability of the feature being discriminative dependent on its empiricial coefficient, which has a tractable sigmoidal form.
\end{theorem}

We provide a proof of this result in \Cref{sec:proofthm}. This finding highlights the effect of the different norms. Optimizing the NADV$_1$ measure assigns infinite costs to all but the dimension that is most likely to be discriminative (with the highest absolute coefficient). On the other hand, the NADV$_\infty$ measure is maximized if the discriminative features exhibit the maximum change of all features, disregarding changes in non-discriminative features. Therefore, the solution attempts to change all dimensions equally through assigning more discriminative dimensions a proportionally higher cost. This ensures that the less discriminative dimensions are altered as well. We observe that $p=2$ seems to constitute a suitable trade-off, where dimensions with low probabilities of being discriminative ($\vp_{\text{disc}}(\hat{\beta}_i)\approx 0$) are penalized by high costs, but the changes will otherwise be distributed evenly among the remaining dimensions.

\section{Experimental Evaluation}

\begin{figure*}[t]
\begin{subfigure}{0.24\textwidth}
\centering
\includegraphics[width=\linewidth]{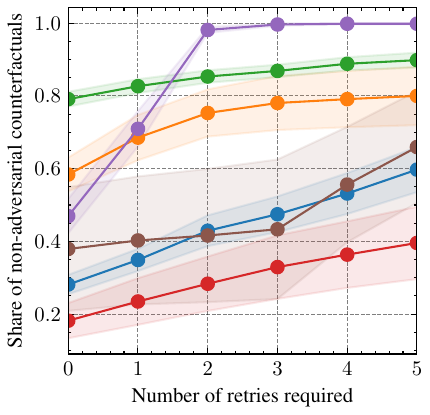}
\vspace{-0.3cm}
\label{fig:retries_admission}
\end{subfigure}
\begin{subfigure}{0.24\textwidth}
\centering
\includegraphics[width=\linewidth]{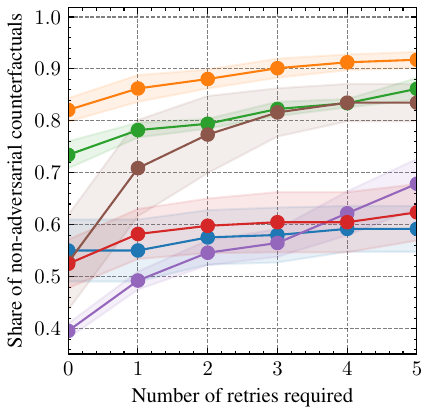}
\vspace{-0.3cm}
\label{fig:retries_german}
\end{subfigure}
\begin{subfigure}{0.24\textwidth}
\centering
\includegraphics[width=\linewidth]{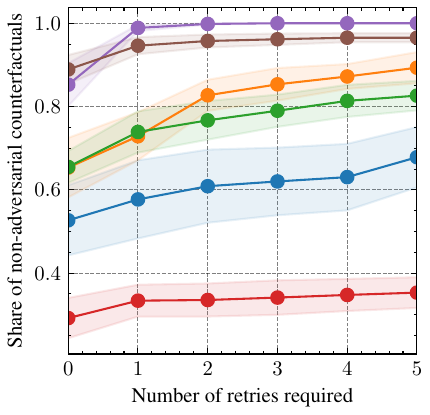}
\vspace{-0.3cm}
\label{fig:retries_compas}
\end{subfigure}
\begin{subfigure}{0.24\textwidth}
\centering
\includegraphics[width=\linewidth]{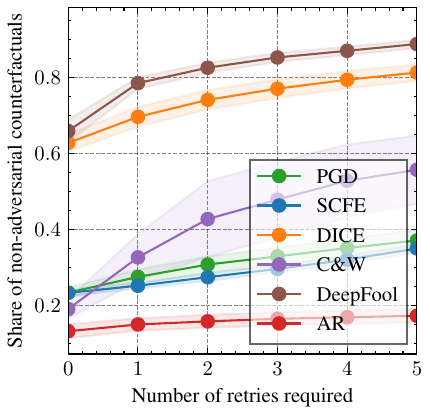}
\vspace{-0.3cm}
\label{fig:retries_heloc}
\end{subfigure}
\begin{subfigure}{0.245\textwidth}
\centering
\includegraphics[width=\linewidth]{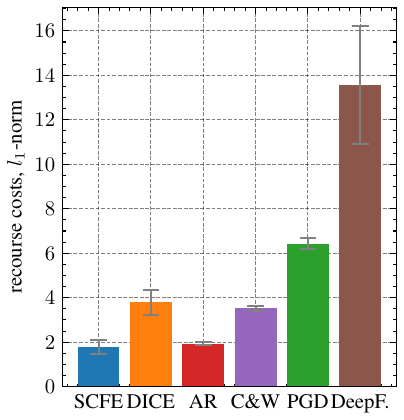}
\vspace{-0.3cm}
\caption{Admission}
\label{fig:retriesc_admission}
\end{subfigure}
\begin{subfigure}{0.245\textwidth}
\centering
\includegraphics[width=\linewidth]{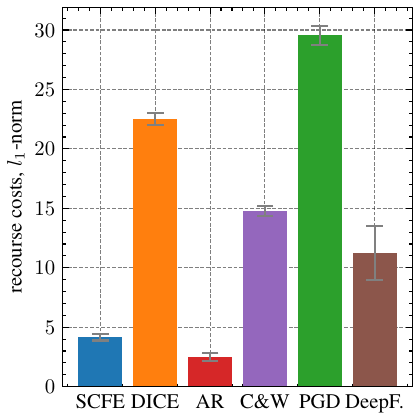}
\vspace{-0.3cm}
\caption{German Credit}
\label{fig:retriesc_german}
\end{subfigure}
\begin{subfigure}{0.245\textwidth}
\centering
\includegraphics[width=\linewidth]{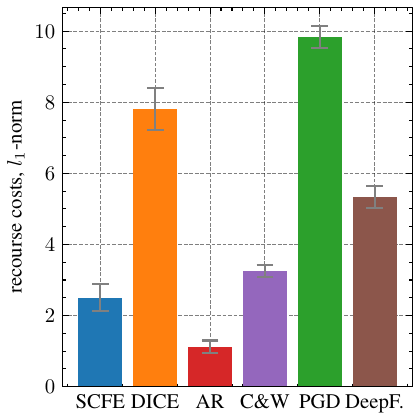}
\vspace{-0.3cm}
\caption{COMPAS}
\label{fig:retriesc_compas}
\end{subfigure}
\begin{subfigure}{0.245\textwidth}
\centering
\includegraphics[width=\linewidth]{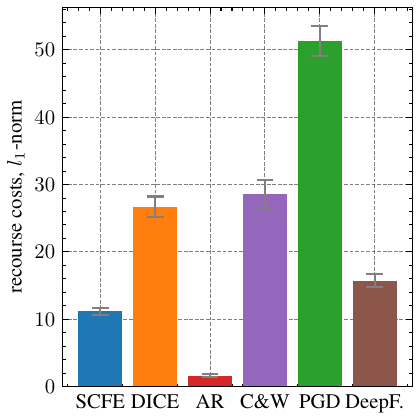}
\vspace{-0.3cm}
\caption{HELOC}
\label{fig:retriesc_heloc}
\end{subfigure}
\caption{\textbf{Both adversarial and recourse methods can succeed in producing non-adversarial recourse for ANNs}. As it might not always be possible to change the ground truth immediately, we study the share of non-adversarial recourse instances after taking a certain number of retries $r$ (a higher share is better). We experiment with three recourse methods (SCFE, DICE, AR) and three adversarial methods (C\&W, PGD, DeepFool). Our results indicate that DICE and PGD usually perform best in identifying non-adversarial counterfactuals. The other adversarial methods, C\&W and DeepFool, often outperform the standard recourse method SCFE regarding non-adversarial recourse. Note that recourse methods strictly optimize for the lowest costs and are therefore less robust than adversarial methods, which incur higher costs.\label{fig:recoursemethods}}
\end{figure*}

\subsection{Experimental Setup}

\textbf{Datasets and  Preprocessing.} To link to the scenario considered in the introduction, we consider four tabular datasets concerned with high-stakes decision-making scenarios where human oversight may be required.

The Law School Admission data set\footnote{\url{https://github.com/mkusner/counterfactual-fairness}} (``admission'') contains information on students from law schools across the United States. Features are collected prior to their entry to law school and include race, sex, entrance exam scores (LSAT), grade-point average (GPA), and regional group. The predicted variable is the z-score of the first-year average grade (ZFYA). The German Credit dataset (``german'') is taken from the UCI machine learning repository\footnote{\url{http://archive.ics.uci.edu/dataset/144/statlog+german+credit+data}} and is concerned with credit scoring. It contains the personal data of 1000 individuals with a binary indicator named ``credit risk'' that serves as a prediction target. The Home Equity Line of Credit (``HELOC'') data set\footnote{\url{https://community.fico.com/s/explainable-machine-learning-challenge}} is a large collection of HELOC applications from anonymized homeowners collected by the financial services provider FICO. The target variable RiskPerformance is ``Bad" if the applicant was at least 90 days past due within the two years after opening the credit account. 
The COMPAS data set\footnote{\url{https://www.kaggle.com/s/danofer/compass}} was initially collected by ProPublica and contains features describing criminal defendants in Broward County, Florida. It also contains their respective recidivism score provided by the COMPAS algorithm and whether or not they reoffended within the following two years. For our analysis, we only kept features relevant for predicting recidivism within the next two years and dropped irrelevant features such as name, date, sex, and race.

For all datasets, continuous features are standardized. Datasets with continuous labels are used in a binary classification fashion where we 
only predict if the z-score exceeds the population’s median.

\textbf{Machine learning models.} We use standard Artificial Neural Networks (ANN) that reflect the implementation of the \texttt{sklearn} library but are implemented in the \texttt{PyTorch} library to leverage automated differentiation capabilities. We train the ANN model (two fully connected hidden layers of width 30) using stochastic gradient descent with the ADAM optimizer. An overview over implementation parameters is provided in the Appendix. 

\textbf{Adversarial Attacks and Recourse Algorithms.}
We implement three powerful adversarial attacks and three recourse methods to study the problem from a practical perspective. We stick to the methods introduced earlier, which include \texttt{SCFE} \cite{wachter2017counterfactual}, which uses a gradient-based objective to find recourses, \texttt{DICE} \citep{mothilal2020fat} with an extra diversity constraint, and \texttt{AR} \citep{Ustun2019ActionableRI}, which uses a Mixed-Integer-Program (MIP) on a discretized action set. Regarding the adversarial attacks, we use \texttt{C\&W} \cite{carlini2017towards} that finds the minimum perturbation on the factual instance to make it change class, \texttt{PGD} \cite{madry2017towards} that uses projected gradients to engender adversarials, and \texttt{DeepFool} \cite{moosavi2016deepfool} that perturbs the input iteratively until the class changes. We adapt the loss function of each optimization algorithm to reflect Eqn.~\eqref{eqn:recourseproblem} and plug in the different cost functions.

\textbf{Ground Truth.} Unfortunately, the number of instances with labels on real-world data sets is limited, such that the ground truth function $y$ is not explicitly available. We, therefore, rely on a simulated ground truth, which uses a subset of the training data that will not be used for model training or testing. We use this data to construct a $k$ nearest neighbor classifier (with $k=5$) that uses a subset of features to simulate an expert committee relying on discriminative features and deciding by majority vote. We then use this ground truth $y$ to predict the remaining instances of the train set. Subsequently, the actual ML model is trained on the remainder of the data and their predictions, making up tuples of the form $\left(\bm{x}, y(\bm{x})\right)$.

\textbf{Evaluation Measures.} Many recourse (and adversarial) methods are implemented to stop right after the model's boundary is crossed. However, this might not initially lead to the non-adversarial recourse desired in practice, even if the correct discriminative features are manipulated (see \Cref{fig:discfeatureretry} for an illustration). We argue that in the practical use case, an individual would query the oracle (e.g., submit their application to the bank again) after obtaining recourse. If the recourse was ineffective in changing the loan decision, an individual could continue to move in the given direction (e.g., further increase their savings amount) until the loan is eventually awarded. We mimic this setup, by increasing the magnitude of $\bdelta = \bm{x}^\prime - \bm{x}$ by 10\% in each step, thus considering $\bm{x}^\prime_r = \bm{x} + (1.1^{r})\bm{\delta}$ after $r \geq 0$ retries. We additionally consider the canonical recourse costs in the $l_1$ and $l_2$ norm.

\subsection{Choice of Optimization Algorithm}
We first put all six implemented methods to the test and check the adversarialness of their outputs. The results are visualized in \Cref{fig:recoursemethods}. We consider the initial recourse and up to 5 more steps in the initial direction. We observe that DICE and PGD usually perform best in identifying non-adversarial counterfactuals. However, the other adversarial methods, C\&W and DeepFool, also often outperform the classical recourse method SCFE regarding non-adversarial recourse. This underlines that, for tabular data, the methods do not reliably produce adversarials. Indeed, they could be considered as recourse methods as well. However, we observe that the adversarial techniques usually result in higher costs, because returning an optimal solution is not their main concern (it just needs to be ``close'' to the input). In contrast, many recourse methods are designed to provide cost-optimal solutions. Non-adversarial recourse is associated with higher cost, leading us to believe that classical recourse methods may be overly cost sensitive for this purpose. We obtained similar results using L2-costs.

\subsection{Choice of Cost Function}
\begin{figure}[b]
\centering
\begin{subfigure}{0.26\textwidth}
\centering
\includegraphics[width=\linewidth]{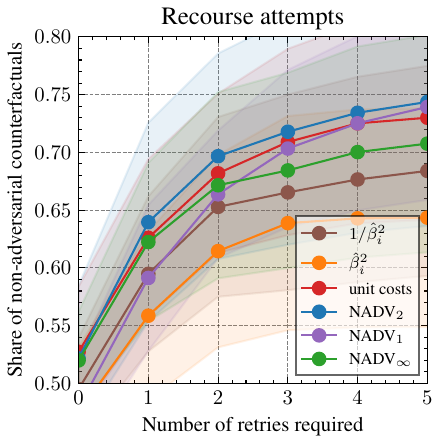}
\vspace{-0.3cm}
\caption{all cost functions}
\label{fig:all_costs_admission}
\end{subfigure}
\begin{subfigure}{0.26\textwidth}
\centering
\includegraphics[width=\linewidth]{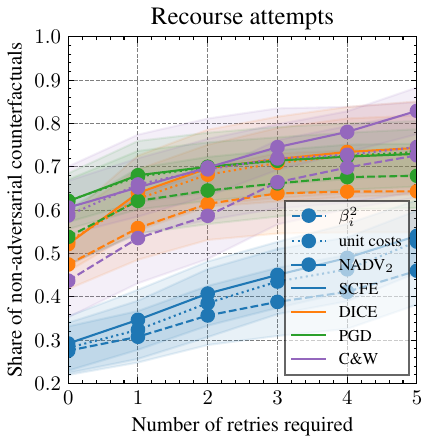}
\vspace{-0.3cm}
\caption{Admission}
\label{fig:costs_admission}
\end{subfigure}
\begin{subfigure}{0.26\textwidth}
\centering
\includegraphics[width=\linewidth]
{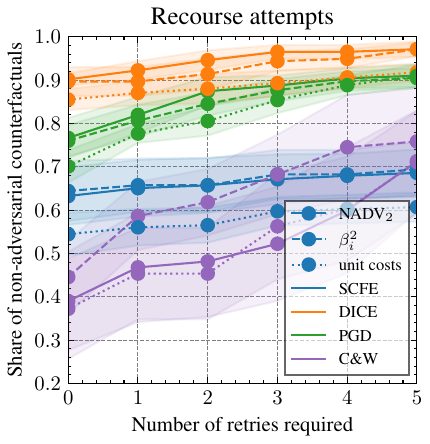}
\vspace{-0.3cm}
\caption{German Credit}
\end{subfigure}
\caption{\textbf{Cost functions can play a role in generating non-adversarial recourse.} (a) ``admission'' dataset with ANN model, DICE results shown. (b,c): Our NADV$_2$ cost function helps in making recourse slightly less adversarial for several methods and thereby reduces the number of retries required.}\label{fig:role_costs}
\vspace{-0.6cm}
\end{figure}

We now study the different cost functions derived in \Cref{sec:optimalcostfunctions} to actual implementations of both recourse and adversarial methods on real data. In particular, we compute the gradients of the model and use the cost weightings derived earlier as well as the default $l_2$-costs, squared gradient costs ($\beta_i^2$, should assign low cost to non-discriminative features) and inverse squared costs ($1/\beta_i^2$) as baselines. DeepFool and AR do not allow for the simple, straightforward inclusion of arbitrary cost functions, so we only consider the four remaining approaches for this experiment and modify their cost-function. The results are shown in \Cref{fig:role_costs}. They show that cost weighting can steer the recourses towards the non-adversarial features and align them better with the ground truth. However, in \Cref{fig:all_costs_admission}, the differences remain statistically insignificant. We observe that the NADV$_2$ optimal weighting scores best among all costs. Inversely weighting the features (e.g., $s_i=\hat{\beta}_i^2$, which assigns low costs to features with almost zero gradients and high costs to features with high gradients), preventing them from being changed, results in the most adversarial recourse. Even though the gap is small, the improvement seems stable across methods (see \Cref{fig:costs_admission}, c) with one exception (C\&W on German Credit). In conclusion, while the cost function can help to make recourse less adversarial, its effect seems to rather subtle.

\subsection{Choice of Machine Learning Model}
\begin{figure}
\begin{subfigure}{0.24\textwidth}
\centering
\includegraphics[width=\linewidth]{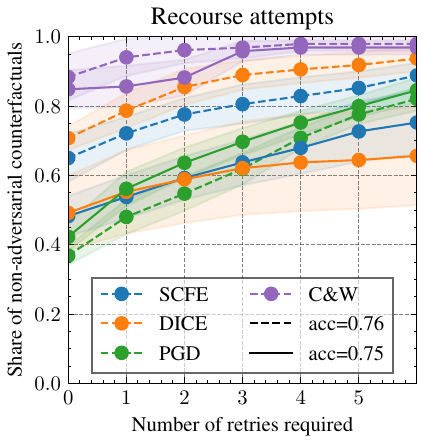}
\vspace{-0.3cm}
\caption{Admission}
\label{fig:accuracy_admission}
\end{subfigure}
\begin{subfigure}{0.24\textwidth}
\centering
\includegraphics[width=\linewidth]{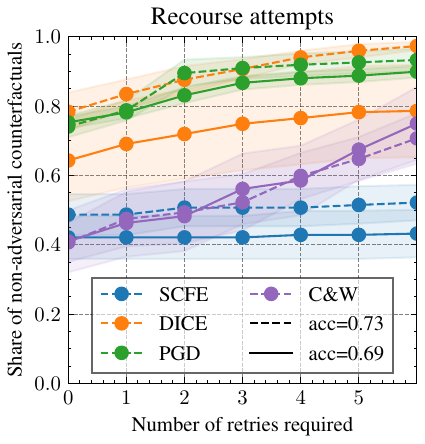}
\vspace{-0.3cm}
\caption{German Credit}
\label{fig:accuracy_german}
\end{subfigure}
\begin{subfigure}{0.24\textwidth}
\centering
\includegraphics[width=\linewidth]{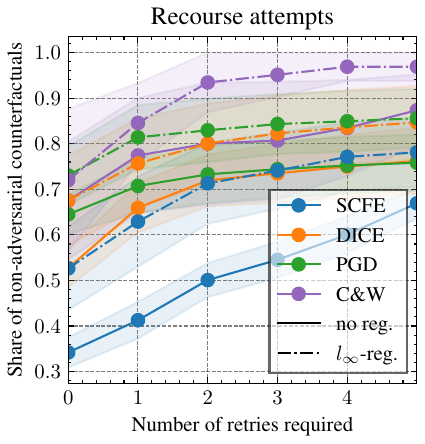}
\vspace{-0.3cm}
\caption{Admission}
\label{fig:protect_admission}
\end{subfigure}
\begin{subfigure}{0.24\textwidth}
\centering
\includegraphics[width=\linewidth]{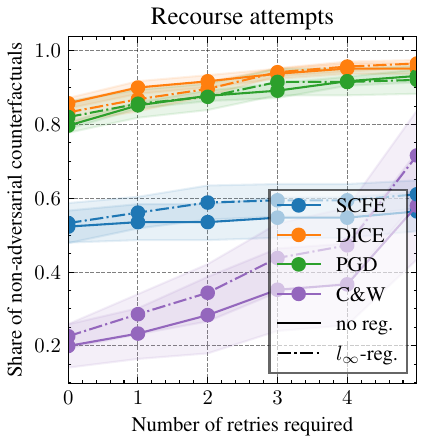}
\vspace{-0.3cm}
\caption{German Credit}
\label{fig:protect_german}
\end{subfigure}

\caption{\textbf{(a,b): More accurate models lead to less adversarial recourse}. We plot the number of retries required to obtain a valid, non-adversarial recourse that changes the ground truth. Logistic Regression Model shown. Results on the remaining datasets can be found in the Appendix. \textbf{(c,d): Regularization through Adversarial Training may improve non-adversarialness}. We robustify models through adversarial training, which improves the share of non-adversarial recourses. \label{fig:role_accuracy}\label{fig:role_adversarialprotect}}
\end{figure}

In our analysis section, we outlined how the machine learning (ML) model may be crucial in determining whether the outcomes can be considered adversarial. We first study the role of the goodness of the model fit. To this end, we train a model on a version of the dataset, where a random sample of 25\% of the data points have flipped labels, which could reflect a realistic use case with noisy human annotations. To rule out other confounding effects to the convexity or smoothness of the model's decision boundary (models trained on noisy labels may have very sharp and more non-convex decision boundaries), we study logistic regression models in this experiment and report the results in  \Cref{fig:role_accuracy}. Surprisingly, the drop in accuracy is not very high (it remains in a range of 1.5\% to 5\%), which we attribute to the datasets being already very noisy previously. Nevertheless, we observe a clear tendency for recourses to be less adversarial for the more accurate models. This trend is stable across datasets and methods.

Adversarial training was proposed by Madry et al.~\cite{madry2017towards} to make models more robust against adversarial attacks. Therefore, it might also offer a suitable way of mitigating adversarial examples in the recourse setup. We study the effect of this form of regularization in an $l_\infty$-ball of radius $\epsilon=0.2$ in \Cref{fig:role_adversarialprotect}. We observe that substantial improvements are possible on the Admission dataset. They are not as pronounced for the remaining datasets but remain visible for most methods. We observe comparable results for the remaining two datasets. Our results highlight that maintaining robust and accurate models one of the most promising strategies towards non-adversarial recourse.


\label{sec:experiments}

\section{Discussion}\label{sec:discussion}


\textbf{Adversarial methods compute recourse on tabular data.} Intriguingly, we observe that despite their purpose, many adversarial attacks succeed in computing non-adversarial recourse on tabular data. While many of the methods were arguably designed with other data modalities, e.g., images, in mind, our finding raises the question of how transferable existing attacks are to variants of the canonical attack scenario. This observation is one in a series of recent claims suggesting that current adversarial attacks may not be realistic in the majority of practical use cases \cite{apruzzese2023real} or require a fundamental paradigm shift away from norms as cost functions towards realistic measures of detector evasion \cite{debenedetti2023evading}.

\textbf{An implicit pursuit towards non-adversarial recourse.} The recourse literature suggests several strategies for improving the quality of recourse. Kommiya et al.~\cite{kommiya2021towards} discovered that feature attributions and feature modifications in recourses only partially agree, raising the question of how they can potentially be used as guidance. Recent takes on robustifying recourse by going further than mandated by the actual decision boundary \cite{pawelczyk2023probabilistically,upadhyay2021robust} can be interpreted as another take to reduce the possibility of ending up with an adversarial. Therefore, we conclude that these works seem to have implicitly followed the goal of obtaining non-adversarial recourse and can be interpreted as orthogonal attempts to reach this common goal. We hope that our precise definition of non-adversarial recourse allows for these efforts to be bundled and unified in the future.

\textbf{Non-Adversarial Recourse via distributional constraints.} Another avenue we have not followed in this work considers the feasible set. The feasible set $\mathcal{X}$ many works have claimed that recourse should be actionable, leading to realistic instances \cite{Ustun2019ActionableRI,poyiadzi2020face}. A fairly general way to arrive at this goal is to constrain the recourse to be in-distribution~\cite{pawelczyk2019,joshi2019towards,dhurandhar2018explanations}, which can be seen as another strategy towards non-adversarial recourse: For in-distribution examples, every model that is a suitable approximation of the ground truth should result in an above-chance-level agreement between the model and the ground truth. We leave an investigation of this connection to future work.

\section{Conclusion}
In this work, we explored the nuanced differences between adversarial examples and counterfactual explanations, focusing on real-world high-stakes decision-making processes. For such scenarios, we introduced the desirable concept of non-adversarial recourse, emphasizing that useful counterfactual explanations should not only change the model's prediction but also align with the ground truth in contrast to adversarial examples.

Our theoretical and experimental analyses on multiple real-world datasets illuminate different ways the model parameters can shape the generation of non-adversarial recourse. Our findings suggest that choosing a suitable model that is highly accurate and robust has more impact on whether recourse can be considered adversarial than the choice of the cost function. For tabular data, adversarial methods also succeed in computing suitable recourse. In summary, we provided valuable insights into generating counterfactuals of reduced adversarialness. Hence, this work lays a foundation for developing resilient recourse models and their deployment in realistic decision-making scenarios.\label{sec:conclusion}
%
%
%
\bibliographystyle{splncs04}
\bibliography{xaiworld_main}

\begin{thebibliography}{10}
\providecommand{\url}[1]{\texttt{#1}}
\providecommand{\urlprefix}{URL }
\providecommand{\doi}[1]{https://doi.org/#1}

\bibitem{abdul2018trends}
Abdul, A., Vermeulen, J., Wang, D., Lim, B.Y., Kankanhalli, M.: Trends and
  trajectories for explainable, accountable and intelligible systems: An hci
  research agenda. In: Proceedings of the 2018 CHI conference on human factors
  in computing systems. pp. 1--18 (2018)

\bibitem{abrate2021counterfactual}
Abrate, C., Bonchi, F.: Counterfactual graphs for explainable classification of
  brain networks. In: KDD (2021)

\bibitem{akhtar2018threat}
Akhtar, N., Mian, A.: Threat of adversarial attacks on deep learning in
  computer vision: A survey. Ieee Access  \textbf{6},  14410--14430 (2018)

\bibitem{apruzzese2023real}
Apruzzese, G., Anderson, H.S., Dambra, S., Freeman, D., Pierazzi, F., Roundy,
  K.: “real attackers don't compute gradients”: Bridging the gap between
  adversarial ml research and practice. In: 2023 IEEE Conference on Secure and
  Trustworthy Machine Learning (SaTML). pp. 339--364. IEEE (2023)

\bibitem{baluja2017adversarial}
Baluja, S., Fischer, I.: Adversarial transformation networks: Learning to
  generate adversarial examples. arXiv preprint arXiv:1703.09387  (2017)

\bibitem{browne2020semantics}
Browne, K., Swift, B.: Semantics and explanation: why counterfactual
  explanations produce adversarial examples in deep neural networks. arXiv
  preprint arXiv:2012.10076  (2020)

\bibitem{carlini2017towards}
Carlini, N., Wagner, D.: Towards evaluating the robustness of neural networks.
  In: 2017 ieee symposium on security and privacy (sp). pp. 39--57. IEEE (2017)

\bibitem{carreira2021counterfactual}
Carreira-Perpi{\~n}{\'a}n, M.{\'A}., Hada, S.S.: Counterfactual explanations
  for oblique decision trees: Exact, efficient algorithms. In: Proceedings of
  the AAAI conference on artificial intelligence. vol.~35, pp. 6903--6911
  (2021)

\bibitem{chen2020strategic}
Chen, Y., Wang, J., Liu, Y.: Strategic recourse in linear classification. arXiv
  preprint arXiv:2011.00355  \textbf{236} (2020)

\bibitem{cheng2015antisocial}
Cheng, J., Danescu-Niculescu-Mizil, C., Leskovec, J.: Antisocial behavior in
  online discussion communities. In: Proceedings of the international aaai
  conference on web and social media. vol.~9, pp. 61--70 (2015)

\bibitem{croce2019sparse}
Croce, F., Hein, M.: Sparse and imperceivable adversarial attacks. In:
  Proceedings of the IEEE/CVF International Conference on Computer Vision. pp.
  4724--4732 (2019)

\bibitem{de2020regression}
De, A., Koley, P., Ganguly, N., Gomez-Rodriguez, M.: Regression under human
  assistance. In: Proceedings of the AAAI Conference on Artificial
  Intelligence. vol.~34, pp. 2611--2620 (2020)

\bibitem{de2021classification}
De, A., Okati, N., Zarezade, A., Rodriguez, M.G.: Classification under human
  assistance. In: Proceedings of the AAAI Conference on Artificial
  Intelligence. vol.~35, pp. 5905--5913 (2021)

\bibitem{debenedetti2023evading}
Debenedetti, E., Carlini, N., Tram{\`e}r, F.: Evading black-box classifiers
  without breaking eggs. arXiv preprint arXiv:2306.02895  (2023)

\bibitem{demir2018patches}
Demir, U., {\"{U}}nal, G.B.: Patch-based image inpainting with generative
  adversarial networks. CoRR  \textbf{abs/1803.07422} (2018),
  \url{http://arxiv.org/abs/1803.07422}

\bibitem{dhurandhar2018explanations}
Dhurandhar, A., Chen, P.Y., Luss, R., Tu, C.C., Ting, P., Shanmugam, K., Das,
  P.: Explanations based on the missing: Towards contrastive explanations with
  pertinent negatives. Advances in neural information processing systems
  \textbf{31} (2018)

\bibitem{dominguez-olmedo22a}
Dominguez-Olmedo, R., Karimi, A.H., Sch{\"o}lkopf, B.: On the adversarial
  robustness of causal algorithmic recourse. In: Proceedings of the 39th
  International Conference on Machine Learning. Proceedings of Machine Learning
  Research, vol.~162, pp. 5324--5342. PMLR (2022)

\bibitem{Du_2022_WACV}
Du, A., Chen, B., Chin, T.J., Law, Y.W., Sasdelli, M., Rajasegaran, R.,
  Campbell, D.: Physical adversarial attacks on an aerial imagery object
  detector. In: Proceedings of the IEEE/CVF Winter Conference on Applications
  of Computer Vision (WACV). pp. 1796--1806 (January 2022)

\bibitem{Duan_2021_CVPR}
Duan, R., Mao, X., Qin, A.K., Chen, Y., Ye, S., He, Y., Yang, Y.: Adversarial
  laser beam: Effective physical-world attack to dnns in a blink. In:
  Proceedings of the IEEE/CVF Conference on Computer Vision and Pattern
  Recognition (CVPR). pp. 16062--16071 (June 2021)

\bibitem{DBLP:conf/iui/FerreiraM21}
Ferreira, J.J., de~Souza~Monteiro, M.: The human-ai relationship in
  decision-making: {AI} explanation to support people on justifying their
  decisions. In: Joint Proceedings of the {ACM} {IUI} 2021 Workshops. vol.~2903
  (2021)

\bibitem{freiesleben2020counterfactual}
Freiesleben, T.: Counterfactual explanations \& adversarial examples--common
  grounds, essential differences, and potential transfers. arXiv preprint
  arXiv:2009.05487  (2020)

\bibitem{freiesleben2022intriguing}
Freiesleben, T.: The intriguing relation between counterfactual explanations
  and adversarial examples. Minds and Machines  \textbf{32}(1),  77--109 (2022)

\bibitem{garcia2017hey}
Garcia, L., Brasser, F., Cintuglu, M.H., Sadeghi, A.R., Mohammed, O.A., Zonouz,
  S.A.: Hey, my malware knows physics! attacking plcs with physical model aware
  rootkit. In: NDSS. pp. 1--15 (2017)

\bibitem{regulation2016regulation}
GDPR: Regulation (eu) 2016/679 of the european parliament and of the council.
  Official Journal of the European Union  (2016)

\bibitem{goodfellow2014explaining}
Goodfellow, I.J., Shlens, J., Szegedy, C.: Explaining and harnessing
  adversarial examples. arXiv preprint arXiv:1412.6572  (2014)

\bibitem{Grudin_2009}
Grudin, J.: Ai and hci: Two fields divided by a common focus. AI Magazine
  \textbf{30}(4), ~48 (Sep 2009). \doi{10.1609/aimag.v30i4.2271},
  \url{https://ojs.aaai.org/aimagazine/index.php/aimagazine/article/view/2271}

\bibitem{guidotti2022counterfactual}
Guidotti, R.: Counterfactual explanations and how to find them: literature
  review and benchmarking. Data Mining and Knowledge Discovery pp. 1--55 (2022)

\bibitem{heath1993induction}
Heath, D., Kasif, S., Salzberg, S.: Induction of oblique decision trees. In:
  IJCAI. vol.~1993, pp. 1002--1007. Citeseer (1993)

\bibitem{ilkhechi2020deepsqueeze}
Ilkhechi, A., Crotty, A., Galakatos, A., Mao, Y., Fan, G., Shi, X., Cetintemel,
  U.: Deepsqueeze: deep semantic compression for tabular data. In: Proceedings
  of the 2020 ACM SIGMOD international conference on management of data. pp.
  1733--1746 (2020)

\bibitem{joshi2019towards}
Joshi, S., Koyejo, O., Vijitbenjaronk, W., Kim, B., Ghosh, J.: Towards
  realistic individual recourse and actionable explanations in black-box
  decision making systems. arXiv preprint arXiv:1907.09615  (2019)

\bibitem{karras2020analyzing}
Karras, T., Laine, S., Aittala, M., Hellsten, J., Lehtinen, J., Aila, T.:
  Analyzing and improving the image quality of stylegan. In: Proceedings of the
  IEEE/CVF conference on computer vision and pattern recognition. pp.
  8110--8119 (2020)

\bibitem{kigma2015adam}
Kingma, D.P., Ba, J.: Adam: {A} method for stochastic optimization. In: Bengio,
  Y., LeCun, Y. (eds.) 3rd International Conference on Learning
  Representations, {ICLR} 2015, San Diego, CA, USA, May 7-9, 2015, Conference
  Track Proceedings (2015), \url{http://arxiv.org/abs/1412.6980}

\bibitem{kommiya2021towards}
Kommiya~Mothilal, R., Mahajan, D., Tan, C., Sharma, A.: Towards unifying
  feature attribution and counterfactual explanations: Different means to the
  same end. In: Proceedings of the 2021 AAAI/ACM Conference on AI, Ethics, and
  Society. pp. 652--663 (2021)

\bibitem{konig2023improvement}
K{\"o}nig, G., Freiesleben, T., Grosse-Wentrup, M.: Improvement-focused causal
  recourse (icr). AAAI Conference on Artificial Intelligence  (2023)

\bibitem{kurakin2016adversarial}
Kurakin, A., Goodfellow, I., Bengio, S., et~al.: Adversarial examples in the
  physical world (2016)

\bibitem{laugel2019dangers}
Laugel, T., Lesot, M.J., Marsala, C., Renard, X., Detyniecki, M.: The dangers
  of post-hoc interpretability: Unjustified counterfactual explanations.
  Proceedings of the Twenty-Eighth International Joint Conference on Artificial
  Intelligence (IJCAI-19)  (2019)

\bibitem{ma2022clear}
Ma, J., Guo, R., Mishra, S., Zhang, A., Li, J.: Clear: Generative
  counterfactual explanations on graphs. Advances in Neural Information
  Processing Systems  \textbf{35},  25895--25907 (2022)

\bibitem{madry2017towards}
Madry, A., Makelov, A., Schmidt, L., Tsipras, D., Vladu, A.: Towards deep
  learning models resistant to adversarial attacks. arXiv preprint
  arXiv:1706.06083  (2017)

\bibitem{moosavi2017universal}
Moosavi-Dezfooli, S.M., Fawzi, A., Fawzi, O., Frossard, P.: Universal
  adversarial perturbations. In: Proceedings of the IEEE conference on computer
  vision and pattern recognition. pp. 1765--1773 (2017)

\bibitem{moosavi2016deepfool}
Moosavi-Dezfooli, S.M., Fawzi, A., Frossard, P.: Deepfool: a simple and
  accurate method to fool deep neural networks. In: Proceedings of the IEEE
  conference on computer vision and pattern recognition. pp. 2574--2582 (2016)

\bibitem{mothilal2020explaining}
Mothilal, R.K., Sharma, A., Tan, C.: Explaining machine learning classifiers
  through diverse counterfactual explanations. In: Proceedings of the 2020
  conference on fairness, accountability, and transparency. pp. 607--617 (2020)

\bibitem{mothilal2020fat}
Mothilal, R.K., Sharma, A., Tan, C.: Explaining machine learning classifiers
  through diverse counterfactual explanations. In: Proceedings of the
  Conference on Fairness, Accountability, and Transparency (FAT*) (2020)

\bibitem{pmlr-v119-mozannar20b}
Mozannar, H., Sontag, D.: Consistent estimators for learning to defer to an
  expert. In: Proceedings of the 37th International Conference on Machine
  Learning. vol.~119, pp. 7076--7087 (2020)

\bibitem{narodytska2016simple}
Narodytska, N., Kasiviswanathan, S.P.: Simple black-box adversarial
  perturbations for deep networks. arXiv preprint arXiv:1612.06299  (2016)

\bibitem{pauwels2023protect}
Pauwels, E.: How to protect biotechnology and biosecurity from adversarial ai
  attacks? a global governance perspective. In: Cyberbiosecurity, pp. 173--184.
  Springer (2023)

\bibitem{pawelczyk2022exploring}
Pawelczyk, M., Agarwal, C., Joshi, S., Upadhyay, S., Lakkaraju, H.: Exploring
  counterfactual explanations through the lens of adversarial examples: A
  theoretical and empirical analysis. In: International Conference on
  Artificial Intelligence and Statistics (AISTATS). pp. 4574--4594. PMLR (2022)

\bibitem{pawelczyk2019}
Pawelczyk, M., Broelemann, K., Kasneci, G.: Learning model-agnostic
  counterfactual explanations for tabular data. In: Proceedings of The Web
  Conference 2020 (WWW). ACM (2020)

\bibitem{pawelczyk2023probabilistically}
Pawelczyk, M., Datta, T., den Heuvel, J.V., Kasneci, G., Lakkaraju, H.:
  Probabilistically robust recourse: Navigating the trade-offs between costs
  and robustness in algorithmic recourse. In: The Eleventh International
  Conference on Learning Representations (ICLR) (2023)

\bibitem{pawelczyk2023on}
Pawelczyk, M., Leemann, T., Biega, A., Kasneci, G.: On the trade-off between
  actionable explanations and the right to be forgotten. In: The Eleventh
  International Conference on Learning Representations (ICLR) (2023)

\bibitem{poyiadzi2020face}
Poyiadzi, R., Sokol, K., Santos-Rodriguez, R., De~Bie, T., Flach, P.: Face:
  feasible and actionable counterfactual explanations. In: Proceedings of the
  AAAI/ACM Conference on AI, Ethics, and Society. pp. 344--350 (2020)

\bibitem{pradel2018deepbugs}
Pradel, M., Sen, K.: Deepbugs: A learning approach to name-based bug detection.
  Proceedings of the ACM on Programming Languages  \textbf{2}(OOPSLA),  1--25
  (2018)

\bibitem{prado2023survey}
Prado-Romero, M.A., Prenkaj, B., Stilo, G., Giannotti, F.: A survey on graph
  counterfactual explanations: Definitions, methods, evaluation, and research
  challenges. ACM Computing Surveys  (2023)

\bibitem{raghu2019algorithmic}
Raghu, M., Blumer, K., Corrado, G., Kleinberg, J., Obermeyer, Z., Mullainathan,
  S.: The algorithmic automation problem: Prediction, triage, and human effort.
  arXiv preprint arXiv:1903.12220  (2019)

\bibitem{rawal2020beyond}
Rawal, K., Lakkaraju, H.: Beyond individualized recourse: Interpretable and
  interactive summaries of actionable recourses. Advances in Neural Information
  Processing Systems  \textbf{33},  12187--12198 (2020)

\bibitem{sahil2010counterfactual}
Sahil, V., Dickerson, J., Hines, K.: Counterfactual explanations for machine
  learning: A review (2010)

\bibitem{stutz2019disentangling}
Stutz, D., Hein, M., Schiele, B.: Disentangling adversarial robustness and
  generalization. In: Proceedings of the IEEE/CVF Conference on Computer Vision
  and Pattern Recognition. pp. 6976--6987 (2019)

\bibitem{su2019one}
Su, J., Vargas, D.V., Sakurai, K.: One pixel attack for fooling deep neural
  networks. IEEE Transactions on Evolutionary Computation  \textbf{23}(5),
  828--841 (2019)

\bibitem{szegedy2013intriguing}
Szegedy, C., Zaremba, W., Sutskever, I., Bruna, J., Erhan, D., Goodfellow, I.,
  Fergus, R.: Intriguing properties of neural networks. arXiv preprint
  arXiv:1312.6199  (2013)

\bibitem{topol2019high}
Topol, E.J.: High-performance medicine: the convergence of human and artificial
  intelligence. Nature medicine  \textbf{25}(1),  44--56 (2019)

\bibitem{upadhyay2021robust}
Upadhyay, S., Joshi, S., Lakkaraju, H.: Towards robust and reliable algorithmic
  recourse. In: Advances in Neural Information Processing Systems (NeurIPS).
  vol.~34 (2021)

\bibitem{Ustun2019ActionableRI}
Ustun, B., Spangher, A., Liu, Y.: Actionable recourse in linear classification.
  In: Proceedings of the Conference on Fairness, Accountability, and
  Transparency (FAT*) (2019)

\bibitem{Ustun_2019}
Ustun, B., Spangher, A., Liu, Y.: Actionable recourse in linear classification.
  Proceedings of the Conference on Fairness, Accountability, and Transparency
  (Jan 2019). \doi{10.1145/3287560.3287566}

\bibitem{verma2020counterfactual}
Verma, S., Dickerson, J., Hines, K.: Counterfactual explanations for machine
  learning: A review. arXiv:2010.10596  (2020)

\bibitem{voigt2017eu}
Voigt, P., Von~dem Bussche, A.: The eu general data protection regulation
  (gdpr). A Practical Guide, 1st Ed., Cham: Springer International Publishing
  \textbf{10},  3152676 (2017)

\bibitem{wachter2017counterfactual}
Wachter, S., Mittelstadt, B., Russell, C.: Counterfactual explanations without
  opening the black box: automated decisions and the gdpr. Harvard Journal of
  Law \& Technology  \textbf{31}(2) (2018)

\bibitem{zhang2022evaluating}
Zhang, J., Lou, Y., Wang, J., Wu, K., Lu, K., Jia, X.: Evaluating adversarial
  attacks on driving safety in vision-based autonomous vehicles. IEEE Internet
  of Things Journal  \textbf{9}(5),  3443--3456 (2022).
  \doi{10.1109/JIOT.2021.3099164}

\bibitem{zhao2022ap}
Zhao, G., Zhang, M., Liu, J., Li, Y., Wen, J.R.: Ap-gan: Adversarial patch
  attack on content-based image retrieval systems. GeoInformatica pp. 1--31
  (2022)

\end{thebibliography}
\appendix
\renewcommand{\thesection}{\Alph{section}}

\section{Derivation of Theorem V.I}
\label{sec:proofthm}
This section presents the proof of \Cref{thm:recourseweight} proof. First, we show how the probability of a relevant feature can be easily estimated in linear models. Suppose we have obtained a data matrix $\mX \in \mathbb{R}^{n \times k}$. Then, we can obtain the analytical least-squares solution $\hat{\vbeta} = (\mX^\top\mX)^{-1}\mX^\top \vy$. We can estimate the variance of $\hat{\vbeta}$ to be $\text{Var}[\hat{\vbeta}] = \sigma^2(\mX^\top\mX)^{-1}$. Simplifying through assuming the features in $\vx$ to be independent and of zero-mean, $\mX^\top\mX$ is diagonal and we obtain
\begin{align}
\text{Var}[\hat{\beta}_i] = \frac{\sigma^2}{\sum_{j=1...n}(\vx_j)_i^2}.
\end{align}
This allows to use of the estimated coefficients to estimate the probability of a feature being relevant, $\vp_{\text{disc}}$ through the following derivation:
\begin{align}
\vp_{\text{disc}}(\hat{\beta}_i) & = \vp(i\in \mathcal{F}_{\text{disc}}|\hat{\beta}_i) \\
& = \frac{\vp(\hat{\beta}_i,i\in \mathcal{F}_{\text{disc}})}{\vp(\hat{\beta}_i, i\in \mathcal{F}_{\text{disc}}) + \vp(\hat{\beta}_i, i\notin \mathcal{F}_{\text{disc}})} \\
& = \frac{\vp(\hat{\beta}_i|i\in \mathcal{F}_{\text{disc}})}{\vp(\hat{\beta}_i | i\in \mathcal{F}_{\text{disc}}) + \vp(\hat{\beta}_i| i\notin \mathcal{F}_{\text{disc}})\underset{q}{\underbrace{\frac{\vp(i\notin \mathcal{F}_{\text{disc}})}{\vp(i\in \mathcal{F}_{\text{disc}})}}}}\\
&=\frac{1}{1+\frac{\vp(\hat{\beta}_i | i\in \mathcal{F}_{\text{disc}})}{q \cdot \vp(\hat{\beta}_i | i\notin \mathcal{F}_{\text{disc}})}} 
 \geq \frac{1}{1+\exp\left(\alpha^2-2\alpha |\hat{\beta}_i| -\log q \right)}\\
& = \text{sigmoid}\left(2\alpha |\hat{\beta}_i| -\alpha^2 +\log q \right) .
\end{align}
The above calculation highlights that it is possible to use the coefficients $\hat{\vbeta}$ in the linear model as noisy estimates for assessing whether a feature is discriminative. 

We combine this insight with the optimal recourse found using a specific cost matrix $\mS$. To this end, we leverage the analytical solution to this problem \cite[Lemma 4, Appendix]{chen2020strategic}:
\begin{align}
\vdelta(\mS) = \underset{c}{\underbrace{\frac{f(\vx)-y_t}{\hat{\vbeta}^\top \mS^{-1} \hat{\vbeta}}}} \mS^{-1} \hat{\vbeta}\label{eqn:analyticalrecourse}.
\end{align}
We can then compute the expected value of the measure of non-adversarialness for the recourse that will be found with the corresponding cost function:
\begin{align}
\mathbb{E}_{\hat{\vbeta}}\left[\text{NADV}_p(\mS) \right] &= \mathbb{E}_{\hat{\vbeta}}\left[\frac{\sum_{i\in \mathcal{F}_{\text{disc}}} \lvert\vdelta_i\rvert}{\lVert\vdelta\rVert_p}\right] = \mathbb{E}_{\hat{\vbeta}}\left[\frac{\sum_{i\in \mathcal{F}_{\text{disc}}} \lvert \frac{\hat{\beta}_i}{s_i} \rvert}{\lVert\mS^{-1} \hat{\vbeta}\rVert_p}\right] \\
& = \frac{\sum_{i} p_{\text{disc},i}(\hat{\vbeta})\lvert \frac{\hat{\beta}_i}{s_i} \rvert}{\lVert\mS^{-1} \hat{\vbeta}\rVert_p} = \frac{\vp_{\text{disc}}^\top(\hat{\vbeta}) (\mS^{-1} |\hat{\vbeta}|)}{\lVert\mS^{-1} \hat{\vbeta}\rVert_p} \\
& = \frac{\vp_{\text{disc}}^\top(\hat{\vbeta}) (\mS^{-1} |\hat{\vbeta}|)}{\lVert\mS^{-1} |\hat{\vbeta}|\rVert_p}
\end{align}
Taking the above expression, we can obtain optimal costs for different values of $p$ by solving 
\begin{align}
\argmax \mathbb{E}_{\hat{\vbeta}}\left[\text{NADV}_p(\mS) \right].
\end{align}
Continuing the calculation separately for the most common values $p\in \left\{1,2,\infty\right\}$, we obtain the following cost weights $s_i$ that depend on the estimated  $\hat{\beta}_i$:\\
\vspace{-.8cm}
\begin{table}[!h]
\centering
\resizebox{0.8\linewidth}{!}{%
\begin{tabular}{p{6cm}cc}
\toprule
\parbox{6cm}{\centering $p=1$} & $p=2$ & $p=\infty$  \\
\midrule
implicit &&\\
\parbox{6cm}{\centering $\mS^{-1} |\hat{\vbeta}|= \kappa\ve_{\argmax_i \vp_{\text{disc}}(\hat{\beta}_i)}$} & $\mS^{-1} |\hat{\vbeta}|= \kappa\frac{\vp_{\text{disc}}(\hat{\vbeta})}{\lVert\vp_{\text{disc}}(\hat{\vbeta})\rVert_2}$ & $\mS^{-1} |\hat{\vbeta}|= \kappa\mathbf{1}$ \\
explicit && \\
\parbox{6cm}{$s_i \sim \left\{1, \text{~if~}i{=}\argmax_j \vp_{\text{disc}}(\hat{\beta}_j), \text{~else~} \infty \right\}$} & $s_i \sim \frac{|\hat{\beta}_i|}{\vp_{\text{disc}}(\hat{\beta}_i)}$ &  $s_i \sim |\hat{\beta}_i|$\\
\bottomrule
\end{tabular}%
}
\end{table}
\vspace{-.9cm}

\begin{table}[hb]
\centering
\resizebox{0.75\linewidth}{!}{%
\begin{tabular}{@{}llll@{}}
\toprule
\multicolumn{2}{l}{}                                          & Artificial Neural Network & Logistic Regression \\ \midrule
\multirow{3}{*}{\rotatebox{90}{Config.}}     & Units     & [Input dim., 30, 30, 2]   & [Input dim., 1]   \\
 & Intermediate activations & ReLU            & N/A                \\
 & Last layer activations   & None         & Sigmoid            \\ \midrule
\multirow{4}{*}{\rotatebox{90}{Training}}
 & Learning rate            & $10^{-3}$       & N/A                \\
 & Regularization           & None            & $l_2$ with pen = 1 \\
 & Batch size & 32 & N/A\\
 & Epochs                   & $10^3 $         & $5 \times 10^3$    \\ \bottomrule
\end{tabular}%
}
\vspace{0.1cm}

\resizebox{\linewidth}{!}{%
\begin{tabular}{lccccr}
    \toprule
    Method & Optimizer & lr & Iterations & $\lambda$  & Additional Comments \\
    \midrule
    {SCFE} & Adam & $10^{-1}$ & $100$ & $0.1$ & $\text{step} = 0$\\
    {DiCE} & RMSProp & $10^{-1}$ & $100$ & - & Two counterfactuals, one is randomly sampled for evaluation \\
    {AR} & Default as in \cite{Ustun2019ActionableRI} & - & - & - &  Squared loss in cost function \\
    {C\&W} & Gradient-based as in \cite{carlini2017towards} & $10^{-2}$ & $1000$ & -  & Constant factor $c = 1$ \\
    {DeepFool} & - & - & $50$ & $2 \times 10^{-2}$  & Target label for attack directionality \cite{moosavi2016deepfool} \\
    {PGD} & - & $10^{-1}$ & $10$ & $10^{-1}$ & $\alpha=10^{-1}$, $\varepsilon = 2$ \\
    \bottomrule
  \end{tabular}
}
\caption{Implementation details.}
\label{tab:classification_models}
\end{table}

\newpage

\section{Implementation Details}\label{sec:impl_details}
\subsection{Preprocessing}
For all datasets, continuous features are standardized. All datasets have predefined train-test splits available online.
We rely on a standard scaler to transform the target variable. We use this dataset in a binary classification fashion and only predict if the z-score exceeds the population's median.
The training set is further split into two parts: i.e., the first is used to train the set of experts, and the second is for the classification models for which adversarial examples and counterfactual explanations are generated. This way, we simulate the experts' knowledge of a small part of the training data on specific actionable features (see \Cref{tab:actionable_features}) and use the rest to learn the classification decision boundary. Adversarial examples and counterfactual explanations are generated for all samples in the test set for a specific classification model.

\subsection{Adversarial and Counterfactual Methods}
Adversarial examples and counterfactuals are generated using the algorithmic implementation described in the original papers. In more detail (see Table \ref{tab:classification_models}),
\begin{itemize}
    \item \texttt{SCFE}: As suggested in \cite{wachter2017counterfactual}, we use an Adam optimizer \cite{kigma2015adam} to obtain the counterfactual explanations. We use $100$ iterations, a learning rate of $10^{-1}$, $\lambda = 0.1$, and we impede the search over the $\lambda$ by setting the step parameter to $0$.
    \item \texttt{DiCE}: We use the RMSProp optimizer with a learning rate of $10^{-1}$ to optimize the proximity and diversity loss that drive the engendering of multiple counterfactuals. We used a maximum number of iterations of $100$ to search for counterfactuals. We compute two counterfactuals, one randomly sampled for the evaluation.
    \item \texttt{AR}: We use the method reported in \cite{Ustun2019ActionableRI} with default parameters. We modify the cost function to incorporate the squared loss used by the other methods.
    \item \texttt{C\&W}: As described in \cite{carlini2017towards}, we use gradient-based optimization to find adversarial examples. We use $1000$ steps to find the adversarials, a $10^{-2}$ learning rate, and a constant factor $c = 1$.
    \item \texttt{DeepFool}: We modify the original implementation \cite{moosavi2016deepfool} to consider a target label to induce the directionality of the attack. We use $50$ steps with an overshoot factor of $2 \times 10^{-2}$. 
    \item \texttt{PGD}: We extend the original method \cite{madry2017towards} to consider different costs - i.e., both components of \Cref{eqn:recourseproblem} - that guide the gradient computation. We use $10$ steps to find the adversarials, $\lambda = 10^{-1}$, $\alpha=10^{-1}$, and $\varepsilon = 2$.    
\end{itemize}

The architecture and training details for our classification models can be found in Table~\ref{tab:classification_models} as well. All models are trained with the same architectures across the datasets on the portion of the training set that is not reserved for the experts' knowledge.

In this work, we suppose that only a limited number of the features is relevant for the ground truth, given in \Cref{tab:actionable_features}, whereas the classifier trained used more features to simulate noise features not considered in the ground truth. We selected features with high global importance for the decision and seemed to be plausible indicators.

\begin{table*}[!t]
\centering
\resizebox{.75\textwidth}{!}{%
\begin{tabular}{@{}ll@{}c}
\toprule
Dataset   & GT Features     & Tot. features            \\ \hline \midrule
Admission & ugpa, first\_pf & 4        \\ \midrule
German Credit &
  \begin{tabular}[c]{@{}l@{}}status, credit-history, employment-duration,\\ housing, number-credits\end{tabular} & 19 \\ \midrule
COMPAS    & age, two\_year\_recid, priors\_count & 5\\ \midrule
HELOC &
  \begin{tabular}[c]{@{}l@{}}MSinceMostRecentTradeOpen, NumTrades60Ever2DerogPubRec,\\ NumTrades90Ever2DerogPubRec, NumTradesOpeninLast12M,\\ NumInqLast6M, NumInqLast6Mexcl7days,\\ NumRevolvingTradesWBalance, NumInstallTradesWBalance,\\ NumBank2NatlTradesWHighUtilization \end{tabular} & 22 \\
 \bottomrule
\end{tabular}%
}
\caption{Features that classifiers, experts, recourse, and adversarial methods can use for each dataset.}
\label{tab:actionable_features}
\end{table*}


\begin{figure*}[tb]
\centering
\begin{subfigure}{0.24\textwidth}
\centering
\includegraphics[width=\linewidth]{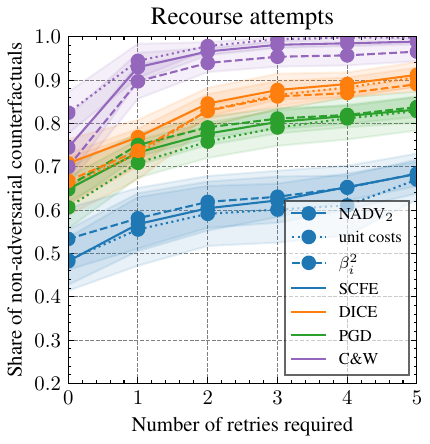}
\vspace{-0.3cm}
\caption{COMPAS}
\end{subfigure}
\begin{subfigure}{0.24\textwidth}
\centering
\includegraphics[width=\linewidth]{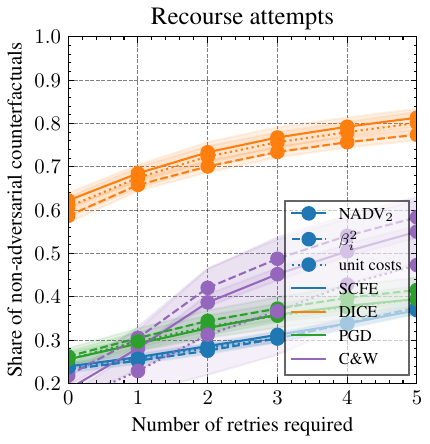}
\vspace{-0.3cm}
\caption{HELOC}
\end{subfigure}

\caption{\textbf{Cost functions can generate non-adversarial recourse but are not highly influential}. We plot the number of required trials to obtain a valid, non-adversarial recourse that changes the ground truth. \label{fig:role_costs_app}}
\vspace{-0.5cm}
\end{figure*} 
\section{Additional Results}\label{sec:additional_results}
This section provides additional plots for more datasets and cost functions.

\begin{figure*}[!t]
\begin{subfigure}{0.24\textwidth}
\centering
\includegraphics[width=\linewidth]{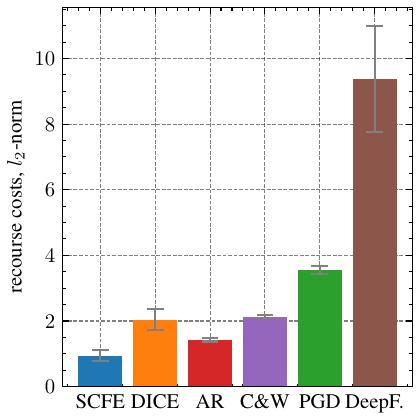}
\vspace{-0.3cm}
\caption{Admission}
\label{fig:retries_admissionl2}
\end{subfigure}
\begin{subfigure}{0.24\textwidth}
\centering
\includegraphics[width=\linewidth]{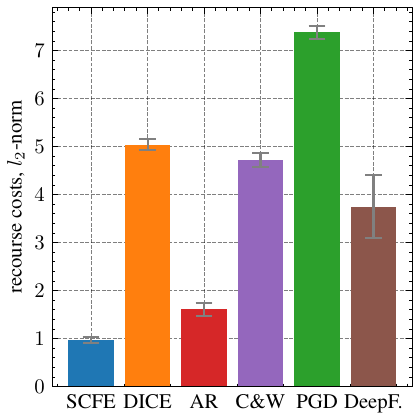}
\vspace{-0.3cm}
\caption{German Credit}
\label{fig:retries_germanl2}
\end{subfigure}
\begin{subfigure}{0.24\textwidth}
\centering
\includegraphics[width=\linewidth]{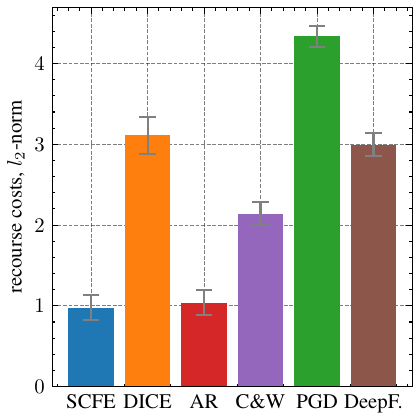}
\vspace{-0.3cm}
\caption{COMPAS}
\label{fig:retries_compasl2}
\end{subfigure}
\begin{subfigure}{0.24\textwidth}
\centering
\includegraphics[width=\linewidth]{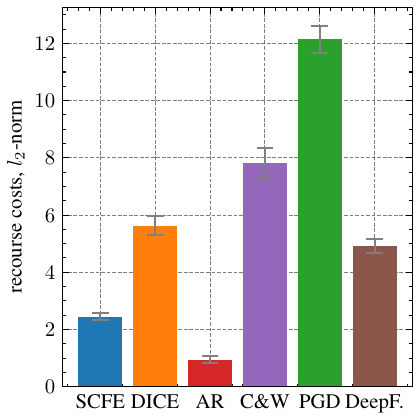}
\vspace{-0.3cm}
\caption{HELOC}
\label{fig:retries_helocl2}
\end{subfigure}
\caption{\textbf{L2 costs for the recourse methods}.\label{fig:l2costs_app}}
\end{figure*}

\textbf{Costs.} We plot the costs of the recourses generated by the methods using the L2 norm in \Cref{fig:l2costs_app}. We observe that the costs in the L2 norm are similar qualitatively to the plot showing L1 costs in the main paper. They confirm our finding that recourse methods SCFE and AR usually lead to lower average costs than adversarial methods.

\textbf{Cost functions.} Results for the remaining datasets can be found in \Cref{fig:role_costs_app}. The NADV$_2$-optimal cost function consistently ranks and usually ranks best except for C\&W on the HELOC and the German Credit dataset, outperforming the unit costs and the inverse weighting, where high-gradient features are assigned higher costs.

\begin{figure}[!h]
\centering
\begin{subfigure}{0.24\textwidth}
\centering
\includegraphics[width=\linewidth]{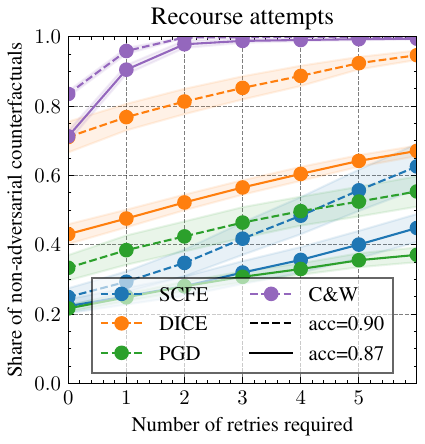}
\vspace{-0.3cm}
\caption{HELOC}
\end{subfigure}
\begin{subfigure}{0.24\textwidth}
\centering
\includegraphics[width=\linewidth]{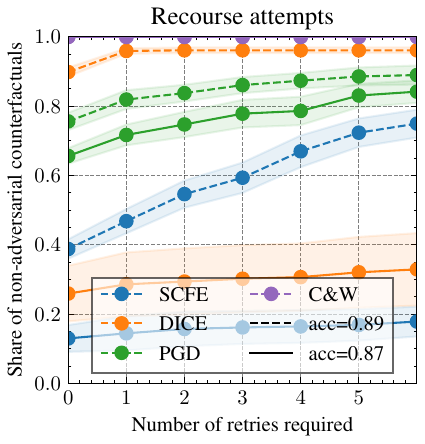}
\vspace{-0.3cm}
\caption{COMPAS}
\end{subfigure}
\caption{\textbf{More accurate models lead to less adversarial recourse}. Results are shown on the COMPAS and HELOC datasets. \label{fig:role_accuracy_app}}
\end{figure}

\textbf{Role of model accuracy.} We illustrate the results for HELOC and COMPAS in \Cref{fig:role_accuracy_app}. They confirm our findings on the other two datasets: i.e.,  more accurate models lead to less adversarial recourse.

\begin{figure}[!t]
\centering
\begin{subfigure}{0.24\textwidth}
\centering
\includegraphics[width=\linewidth]{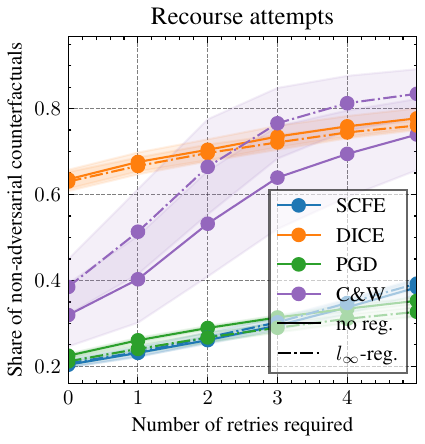}
\vspace{-0.3cm}
\caption{Heloc}
\end{subfigure}
\begin{subfigure}{0.24\textwidth}
\centering
\includegraphics[width=\linewidth]{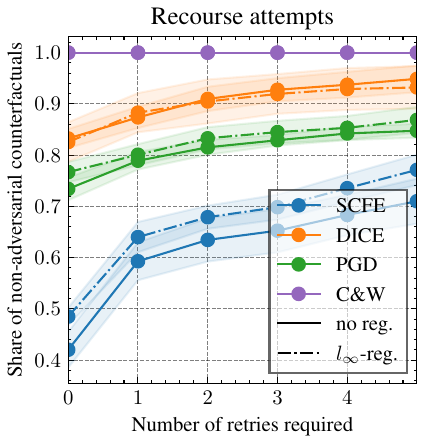}
\vspace{-0.3cm}
\caption{COMPAS}
\end{subfigure}
\caption{\textbf{Regularization through Adversarial Training may improve non-adversarialness}. We robustify models through adversarial training, which seems to improve the share of non-adversarial recourses slightly. \label{fig:role_adversarialprotect_app}}
\end{figure}

\textbf{Regularization through adversarial training.} Results for HELOC and the COMPAS dataset can be found in \Cref{fig:role_adversarialprotect_app}. Notice that the regularized version (full line) tends to improve over the non-regularized baseline (dashed line). Intriguingly, C\&W on COMPAS is perfect at finding non-adversarial counterfactuals without any retries required. This leads us to believe that C\&W is not an effective adversarial attack in this scenario. It rather behaves as a recourse method, which strengthens our motives presented in \Cref{sec:discussion} about the adaptation of adversarial methods to real-world scenarios.


\end{document}